\documentclass[10pt,twocolumn,letterpaper]{article}

\usepackage[pagenumbers]{iccv} 

\usepackage{algorithm}
\usepackage{algorithmic}

\usepackage{amsmath}
\usepackage{amssymb}
\usepackage{mathtools}
\usepackage{amsthm}
\usepackage{multirow}
\usepackage[table,dvipsnames]{xcolor}
\usepackage{marvosym}
\usepackage{listings}
\definecolor{iccvblue}{rgb}{0.21,0.49,0.74}
\usepackage[pagebackref,breaklinks,colorlinks,allcolors=iccvblue]{hyperref}

\usepackage{symbols}

\usepackage{amsmath,amsfonts,bm}

\def\1{\bm{1}}

\def\sp{space}

\def\vv{{\bm{v}}}

\def\mI{{\bm{I}}}

\DeclareMathAlphabet{\mathsfit}{\encodingdefault}{\sfdefault}{m}{sl}
\SetMathAlphabet{\mathsfit}{bold}{\encodingdefault}{\sfdefault}{bx}{n}

\def\gL{{\mathcal{L}}}

\def\gN{{\mathcal{N}}}

\def\sR{{\mathbb{R}}}

\theoremstyle{plain}

\theoremstyle{definition}

\theoremstyle{remark}

\makeatletter
\DeclareRobustCommand{\pdot}{\mathbin{\mathpalette\pdot@\relax}}
\newcommand{\pdot@}[2]{%
  \ooalign{%
    $\m@th#1\circ$\cr
    \hidewidth$\m@th#1\cdot$\hidewidth\cr
  }%
}
\makeatother

\newcommand*{\ShowNotes}{} 
\definecolor{darkred}{rgb}{0.7,0.1,0.1}
\definecolor{darkgreen}{rgb}{0.1,0.7,0.1}
\definecolor{cyan}{rgb}{0.7,0.0,0.7}
\definecolor{dblue}{rgb}{0.2,0.2,0.8}
\definecolor{maroon}{rgb}{0.76,.13,.28}
\definecolor{burntorange}{rgb}{0.81,.33,0}
\definecolor{tealblue}{rgb}{0.212,0.459, 0.533}

\definecolor{mypink}{rgb}{0.93359375, 0.62109375, 0.83984375}

\definecolor{pp}{rgb}{0.43921569, 0.18823529, 0.62745098}
\definecolor{rr}{rgb}{0.5254902 , 0.00784314, 0.12941176}
\definecolor{bb}{rgb}{0.09019608, 0.23529412, 0.37647059}
\definecolor{yy}{rgb}{0.49803922, 0.3372549 , 0.0}
\definecolor{gg}{rgb}{0.02352941, 0.3372549 , 0.17647059}

\ifdefined\ShowNotes
  \newcommand{\colornote}[3]{{\color{#1}\bf{#2: #3}\normalfont}}
\else
  \newcommand{\colornote}[3]{}
\fi

\usepackage{thmtools}
\usepackage{thm-restate}

\newcommand{\ours}[0]{$\text{CFG-Zero$^\star$}$\xspace}

\def\paperID{1012} 
\def\confName{ICCV}
\def\confYear{2025}

\title{
\ours: Improved Classifier-Free Guidance for Flow Matching Models
}

\author{Weichen Fan$^{1}$ \hspace{.2cm}  Amber Yijia Zheng$^{2}$ \hspace{.2cm} Raymond A. Yeh$^{2}$ \hspace{.2cm}  Ziwei Liu$^{1,}$\textsuperscript{\Letter} \\
\textsuperscript{1}S-Lab, Nanyang Technological University  \\
\textsuperscript{2}Department of Computer Science, Purdue University\\
{\tt\small weichen.fan@u.nus.edu, \{zheng709,rayyeh\}@purdue.edu,}\\
{\tt\small ziwei.liu@ntu.edu.sg}}

\begin{document}

\twocolumn[{%
\renewcommand\twocolumn[1][]{#1}%
\maketitle

\begin{center}
    \captionsetup{type=figure}
    \small
    \vspace{-1cm}
\includegraphics[width=0.8\linewidth]{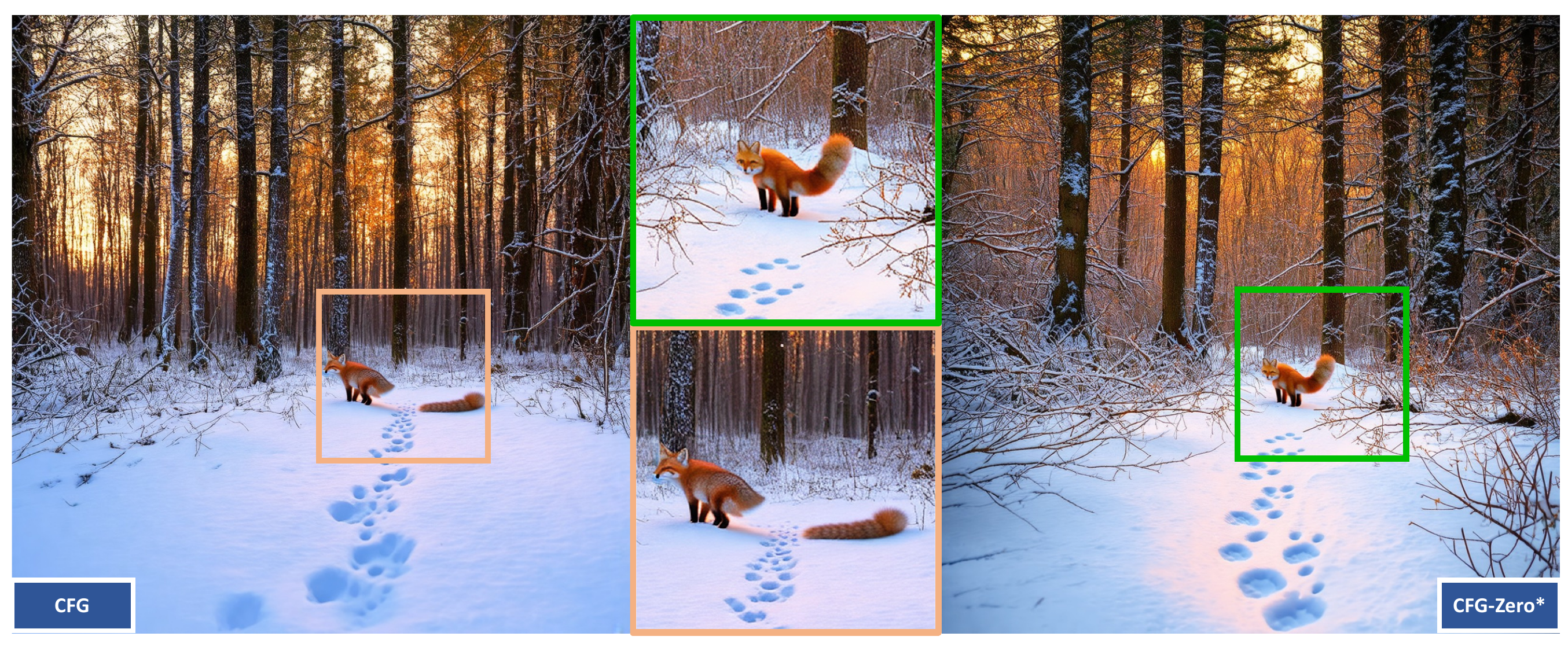}
    \vspace{-3mm}
    \captionof{figure}{Comparison for the prompt: \textit{``A dense winter forest with snow-covered branches, the golden light of dawn filtering through the trees, and a lone fox leaving delicate paw prints in the fresh snow.''} Images generated using SD3.5~\cite{sd3} with CFG and \ours(Ours).
    }
    \label{fig:teaser_img}
\end{center}}]

\begin{abstract}
Classifier-Free Guidance (CFG) is a widely adopted technique in diffusion/flow models to improve image fidelity and controllability. In this work, we first analytically study the effect of CFG on flow matching models trained on Gaussian mixtures where the ground-truth flow can be derived. We observe that in the early stages of training, when the flow estimation is inaccurate, CFG directs samples toward incorrect trajectories. Building on this observation, we propose \textbf{\ours}, an improved CFG with two contributions: 
{\bf (a) optimized scale}, where a scalar is optimized to correct for the inaccuracies in the estimated velocity, hence the $\star$ in the name; and 
{\bf (b) zero-init}, which involves zeroing out the first few steps of the ODE solver.
Experiments on text-to-image (Lumina-Next, Stable Diffusion 3, and Flux) and text-to-video (Wan-2.1) generation demonstrate that \ours consistently outperforms CFG, highlighting its effectiveness in guiding Flow Matching models. (\textbf{Code is available at \url{github.com/WeichenFan/CFG-Zero-star}})

\end{abstract}
\vspace{-5mm}
\section{Introduction}
Diffusion and flow-based models are the state-of-the-art (SOTA) for generating high-quality images and videos, with recent advancements broadly categorized into Score Matching~\cite{ho2020denoising,song2020denoising,song2019generative,song2020score,rombach2022high,peebles2023scalable} and Flow Matching~\cite{liu2022flow,sd3,fan2025vchitect,luminanext,flux} approaches. Flow matching directly predicts a velocity, enabling a more interpretable transport process and faster convergence compared with score-based diffusion methods. Hence, recent SOTA in text-to-image/video models increasingly adopt flow matching. In this paper, we follow the unifying perspective on diffusion and flow models presented by~\citet{lipman2024flow}. We broadly use the term \textit{flow matching models} to refer to any model trained using flow matching, where samples are generated by solving an ordinary differential equation (ODE).

\begin{figure}[t]
\centering
\includegraphics[width=0.95\linewidth]{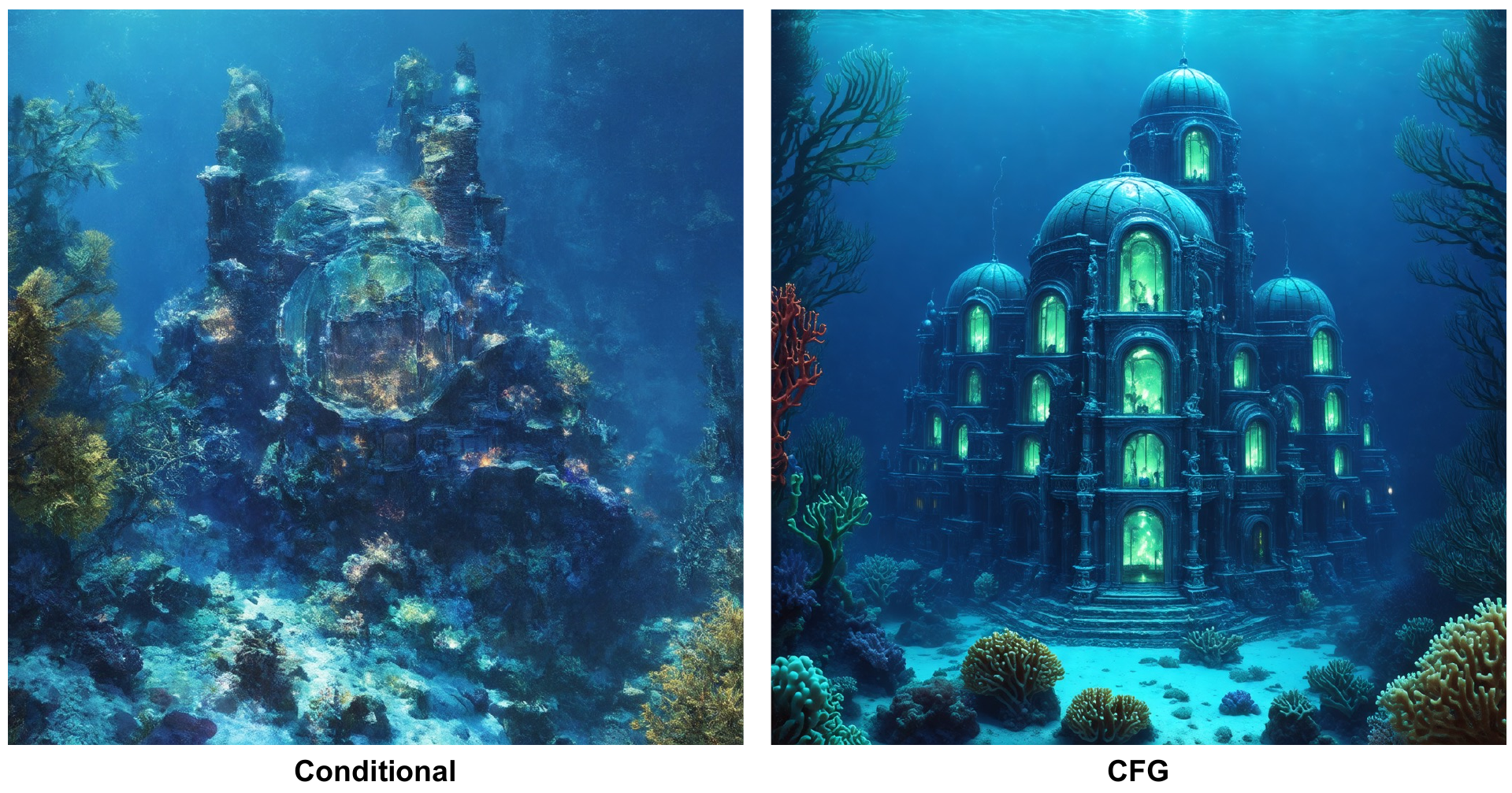}
\caption{\textbf{(Left)} Conditional generation. \textbf{(Right)} CFG generation. (Prompt: \textit{``A mysterious underwater city with bioluminescent corals and towering glass domes.''})}
\label{fig:cond_cfg}
\vspace{-4mm}
\end{figure}

Next, classifier-free guidance (CFG)~\cite{ho2022classifier,zheng2023guided} is a widely used technique in flow matching models to improve sample quality and controllability during generation. In text-to-image tasks, CFG improves the alignment between generated images and input text prompts.  
In other words, CFG is used because the conditional distribution induced by the learned conditional velocity does not fully match with the user's ``intended'' conditional distribution; see example in~\figref{fig:cond_cfg}.%
We hypothesize that this mismatch arises from two fundamental factors. First, it may be from dataset limitations, where the user's interpretation of a text prompt and its corresponding image differs from the dataset distribution. Second, it could result from a learning limitation, where the learned velocity fails to accurately capture the dataset's distribution. In this work, we focus on the latter issue. 
When the model is underfitted, a 
mismatch exists between the conditional and unconditional predictions during sampling, causing CFG to guide the sample in a direction that deviates significantly from the optimal trajectory. Specifically, the velocity estimated by CFG in the first step at $x_0$ may contradict the optimal velocity. This suggests that skipping this prediction could lead to better results.

We empirically analyze the effect of CFG when the learned velocity is underfitted,~\ie, inaccurate, using a mixture of Gaussians as the data distribution. In this setting, the ground-truth (optimal) velocity has a closed-form solution, allowing us to compare it with the learned velocity throughout training. 
Based on the observations, we then propose \ours, which introduces two key improvements over the vanilla CFG: \textit{optimized scale} and the \textit{zero init} technique. 

In~\secref{Sec:Method}, we provide an empirical analysis to motivate our approach with supporting observations. Furthermore, we validate that the observations go beyond mixture of Gaussians by conducting experiments on ImageNet for the task of class-conditional generation. Finally, in~\secref{Sec:exp}, we apply \ours to text-guided generation, showing that our method achieves strong empirical performance in guiding SOTA flow matching models.

{\bf \noindent Our main contributions are summarized as follows:}
\begin{itemize}
    \item We analyze the sources of error in flow-matching models and propose a novel approach to mitigate inaccuracies in the predicted velocity.

    \item We empirically show that zeroing out the first ODE solver step for flow matching models improves sample quality when the model is underfitted. 
    
    \item Extensive experiments validate that \ours achieves competitive performance in both discrete and continuous conditional generation tasks, demonstrating its effectiveness as an alternative to CFG.
\end{itemize}

\section{Related Work}

{\noindent\bf Diffusion and Flow-based Models.} Unlike generative adversarial methods~\cite{goodfellow2020generative} that rely on one-step generation, diffusion models~\cite{dhariwal2021diffusion} have demonstrated significantly improved performance in generating high-quality samples. Early diffusion models were primarily score-based generative models, including DDPM~\cite{ho2020denoising}, DDIM~\cite{song2020denoising}, EDM~\cite{karras2022elucidating}, and Stable Diffusion~\cite{rombach2022high}, which focused on learning the SDEs governing the diffusion process. 

Next, Flow Matching~\cite{lipmanflow} provides an alternative approach by directly modeling sample trajectories using ordinary differential equations (ODEs) instead of SDEs. This enables more stable and efficient generative processes by learning a continuous flow field that smoothly transports samples from a prior distribution to the target distribution. Several works, including Rectified Flow~\cite{liu2022flow}, SD3~\cite{sd3}, Lumina-Next~\cite{luminanext}, Flux~\cite{flux}, Vchitect-2.0~\cite{fan2025vchitect}, Lumina-Video~\cite{luminavideo} HunyuanVideo~\cite{kong2024hunyuanvideo}, SkyReels-v1~\cite{SkyReelsV1}, and Wan2.1~\cite{wan2.1} have demonstrated that ODE-based methods achieve faster convergence and improved controllability in text-to-image and text-to-video generation. As a result, Flow Matching has become a compelling alternative to stochastic diffusion models, offering better interpretability and training stability. Thus, our analysis is based on Flow Matching models, which aim to provide more accurate classifier-free guidance. 

{\noindent\bf Guidance in Diffusion Models.}  
Achieving better control over diffusion models remains challenging yet essential. Early approaches, such as classifier guidance (CG)~\cite{dhariwal2021diffusion}, introduce control by incorporating classifier gradients into the sampling process. However, this method requires separately trained classifiers, making it less flexible and computationally demanding. To overcome these limitations, classifier-free guidance (CFG)~\cite{ho2022classifier} was proposed, enabling guidance without the need for an external classifier. Instead, CFG trains conditional and unconditional models simultaneously and interpolates between their outputs during sampling.

Despite its effectiveness, CFG relies on an unbounded empirical parameter, known as the guidance scale, which determines how strongly the generated output is influenced by the conditional model. Improper tuning of this scale can lead to undesirable artifacts, either over-saturated outputs with excessive conditioning or weakened generation fidelity due to under-conditioning. In fact, previous studies~\cite{bradley2024classifier} have observed that CFG estimation does not provide an optimal denoising direction. 

As a result, several studies have explored adaptive or dynamically scaled guidance to address these issues, including ADG~\cite{sadat2024eliminating}, Characteristic-Guidance~\cite{zheng2023characteristic}, ReCFG~\cite{xia2024rectified}, Weight-Scheduler~\cite{xianalysis}, and CFG++~\cite{chung2024cfgplusplus}. Additionally, AutoG~\cite{karras2024guiding} replaces the unconditional model with a smaller, less-trained version of the model itself. Other researchers~\cite{kynkaanniemi2025applying} have proposed limiting the use of CFG to a specific interval during sampling.

Unlike previous approaches, our work is motivated by an observation that CFG prediction is inaccurate when a model is underfitted, and specifically in the first step, \ie, the prediction is even worse than zeroing out the first ODE solver step. Thus, we 
study the error of CFG within the Flow Matching, where we derive the upper bound of the error term and propose to minimize it. From this analysis, we derive a dynamic parameterization technique that adjusts the unconditional output, leading to more stable and effective guidance in Flow Matching based diffusion models.

\section{Preliminaries}

We briefly recap flow matching following the unifying perspective presented in~\citet{lipman2024flow}.

{\bf\noindent Conditional flow matching.} Given a source distribution $p(x|y)$ and an unknown target distribution $q(x|y)$, conditional flow matching (CFM) defines a probability path $p_t(x|y)$, where $t \in [0, 1]$ is a continuous time variable that interpolates between $p$ and $q$, such that $p_0(x|y) = p(x|y)$ and $p_1(x|y) = q(x|y)$. An effective choice for $p_t(x|y)$ is the linear probability path
\bea
p_t(x|y) \triangleq (1-t) \cdot p(x|y) + t \cdot q(x|y),
\eea
where a sample
$x_t = (1-t) x_0 + t x_1,$
with $x_0 \sim p(x|y)$ and $x_1 \sim q(x|y)$. 
Next, a continuous flow model is trained by learning a time-dependent velocity field $\frac{d}{dt}x_t = \vv_t^\theta(x|y)$ that governs the trajectory of $x$ over $t$. Here, the velocity is represented using a deep-net with trainable parameters $\theta$. A flow matching model is trained by minimizing the CFM loss expressed as
\bea
L_{\text{CFM}}(\theta) = \mathbb{E}_{t, x_0, x_1} \left\| \vv_t^\theta\big(x_t |y\big) - \big(x_1 - x_0\big) \right\|_2^2.
\eea
At generation time, a new sample can be obtained by using any ODESolver,~\eg, the midpoint method~\cite{suli2003introduction}. 

{\bf\noindent Classifier free guidance (CFG)}~\cite{ho2022classifier,zheng2023guided} improves the quality of conditional generation by steering a sample toward the given input condition, \eg, a class label or a text prompt. In CFG, a single flow model $\vv^\theta_t(x| y)$ is trained to output both conditional and unconditional velocity fields. This is done by introducing $y=\emptyset$, which does not contain any conditioning information.

At inference, the guided velocity field is formed by
\bea
\hat{\vv}_t^\theta(x|y) \triangleq (1-w) \cdot \vv^\theta_t(x|y=\emptyset) + w \cdot \vv^\theta_t(x|y),
\eea
where $\omega$ is the guidance scale and $\varnothing$ denotes the null condition. When $\omega=1$, it is equivalent sampling only using the conditional velocity $\vv^\theta_t(x|y)$, \ie, no guidance.

{\noindent\bf  Closed form velocity for Gaussian distributions.}
When both the source and target consist of Gaussian mixtures, the optimal velocity has a closed form.

For example, let the source distribution $p = \gN(0,\mI)$ and target distribution $q = \gN(\mu, \mI)$ both be a single Gaussian. Then, the corresponding optimal velocity $\vv_t^\ast(x) =$
\bea
 \Bigl[(2t-1)\mI\Bigr]\Bigl[(1-t)^2\mI + t^2\mI\Bigr]^{-1} (x - t\mu) + \mu.
\eea
With a closed-form optimal $\vv_t^*$, we can now empirically study the gap between the optimal and learned velocity $\norm{\vv_t^*(\cdot)-\vv_t^\theta(\cdot)}$ throughout training, specifically in the underfiting regime, which motivated our proposed \ours.

\begin{algorithm}[t]
\caption{\ours}
\label{alg:cfg-zero}
\begin{algorithmic}[1]
\STATE \textbf{Input:} Trained velocity $\vv^\theta$, noise sample $x_0$, guidance weight $\omega$, and number of steps to zero out $K$.

{\color{iccvblue} \# $K$ equals to 1 by default}
\STATE $s_t^\star \leftarrow \frac{\vv^\theta_t(\cdot|y)^\top\vv^\theta_t(\cdot|\emptyset)}{\norm{\vv^\theta_t(\cdot|\emptyset)}^2}$ {\color{iccvblue} \# Optimized scale}
\STATE $\tilde{\vv}_t(\cdot) \leftarrow  (1-\omega)\cdot s_t^\star \cdot   \vv^\theta_t(\cdot|\emptyset)) + \omega \cdot \vv^\theta_t(\cdot|y)$

\STATE {\color{iccvblue} \# Solve ODE}
\FOR{$t = 0$ \textbf{to} $T$}
\IF{$t<K$} 
\STATE $x_{t+1} \leftarrow x_t$ {\color{iccvblue} \# Zero-init}
\ELSE
\STATE $x_{t+1} \leftarrow \text{ODEStep}(\tilde{\vv}_t(\cdot), x_t)$
\ENDIF
\ENDFOR
\STATE \textbf{Return} generated sample $x_T$
\end{algorithmic}
\end{algorithm}

\section{Methodology}
\label{Sec:Method}
We propose a guidance algorithm \ours with two improvements from the standard CFG: {\bf (a) optimized scale}, where a scalar parameter is optimized to compensate for inaccuracies in the learned velocity (\secref{sec:opt_scale}); 
{\bf (b) zero-init}, where we zero out the first step of the ODE solver (\secref{sec:zero_init}). These modifications can be easily integrated into the existing CFG code base and introduce minimal additional computational cost. The overall algorithm is summarized in~\algref{alg:cfg-zero}.

\subsection{Optimizing an additional scaler in CFG}\label{sec:opt_scale}

As motivated in the introduction, we aim to study the use of CFG in the setting where the velocity is underfitted, \ie, CFG is used in the hope that the guided velocity $\tilde{v}^\theta$ approximates the ground-truth flow $\tilde{v}^\star$, \ie,
\bea\label{eq:approx}
\tilde{\vv}^\theta_t(x|y) \approx \vv_t^*(x|y).
\eea

To further improve this approximation, we introduce an optimizable scaler $s \in \sR_{>0}$ to CFG,
\bea\label{eq:weight_guidance}
\tilde{\vv}^\theta_t(x|y) \triangleq & (1-\omega) \cdot s \cdot \vv^\theta_t(x) + \omega \cdot \vv^\theta_t(x|y)\\
 =& -\omega'\cdot s \cdot \vv^\theta_t(x) + (1+\omega') \vv^\theta_t(x|y),
\eea
where $\omega' \triangleq 1+\omega$. 
The choice of learning a scaler $s$ is inspired by classifier guidance~\cite{dhariwal2021diffusion}, where they introduce a scaling factor to balance between the gradient and the unconditioned direction. The remaining challenge is how to optimize $s$.

In a hypothetical case where we do have ground-truth flow $\vv_t^*$, then one can formulate the approximation in~\equref{eq:approx} as a least-squares, \ie, minimizing $s$ over the loss
\bea \label{eq:lsq}
\gL(s) \triangleq \norm{\tilde{\vv}^\theta_t(x|y) - \vv_t^*(x|y)}_2^2.
\eea
However, as $\vv_t^*$ is unknown, we instead minimize over an upperbound of $\gL(s)$ in~\equref{eq:lsq} established using triangle inequality as follows:
\bea\nonumber
\gL(s) =& \norm{
-\omega' s \vv^\theta_t(x) + (1+w') \vv^\theta_t(x|y)
- \vv_t^*(x|y)}_2^2\\ \nonumber
\leq& \hspace{-2.75cm}\norm{\vv^\theta_t(x|y)}_2^2 
+ \norm{\vv_t^*(x|y)}_2^2\\
& \hspace{1.22cm} + \; \omega'\norm{\vv^\theta_t(x|y)- s \cdot \vv^\theta_t(x)}_2^2.
\label{eq:upper}
\eea
Observe that only the last term has dependencies on $s$, \ie, optimizing~\equref{eq:upper} is equivalent to

\bea
\min_s \norm{\vv^\theta_t(x|y)-s \cdot \vv^\theta_t(x)}_2^2,
\eea
where the solution $s^\star$ is a projection of the condition velocity onto the unconditional velocity, \ie,
\bea
s^\star = (\vv_t^\theta(x|y)^\top\vv_t^\theta(x))/{\norm{\vv_t^\theta(x)}^2}.
\eea

\begin{figure}[t]
    \centering
    \setlength{\tabcolsep}{0pt}
    \begin{tabular}{cc}
    \includegraphics[width=0.5\linewidth, height=3.19cm]{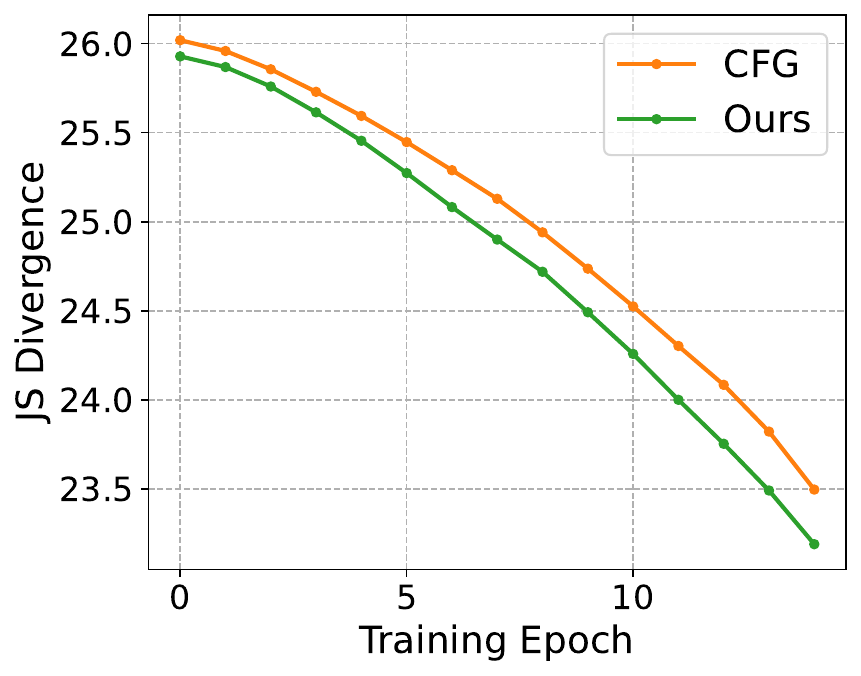} &
    \includegraphics[width=0.5\linewidth,]{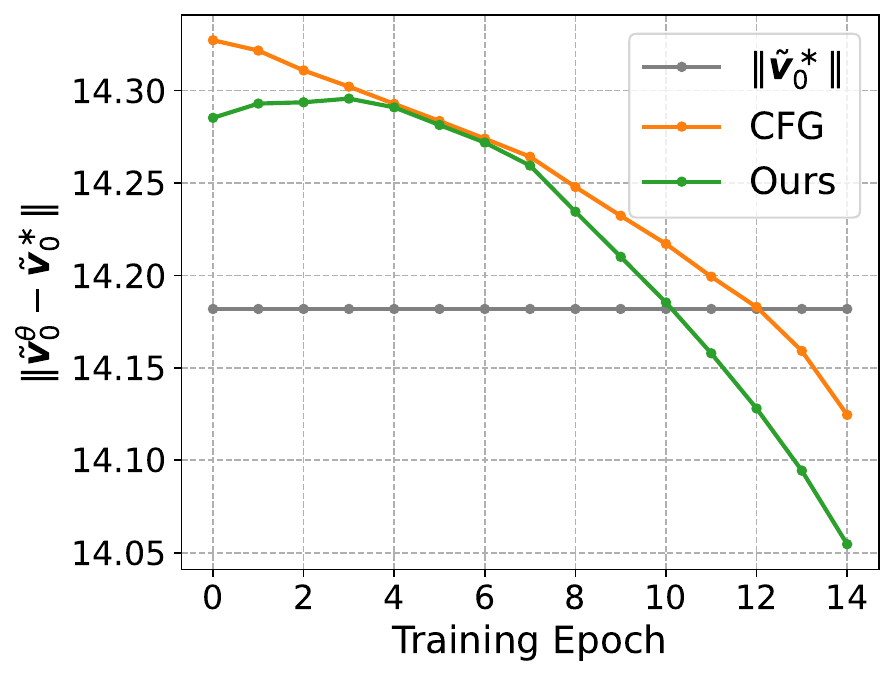}\\
    (a) & (b)
    \end{tabular}
    \vspace{-0.15cm}
    \caption{\textbf{Results on mixture of Gaussians in $\sR^2$.} \textbf{Left:} The Jensen–Shannon divergence between the model's final flow sample distribution and the target distribution \textit{v.s.} training epoch. {\bf Right:} The velocity error norm $\|\tilde{\vv}^\theta_0 - \tilde{\vv}^\ast_0\|$, with the ground truth norm shown in gray \textit{v.s.} training epoch. 
    }
    \vspace{-0.15cm}
    \label{fig:toy_jsd_norm}
\end{figure}

We empirically validate the approach on a toy example consisting of a Gaussian mixture, with results shown in~\figref{fig:toy_jsd_norm}(a), where we observe that samples generated with \ours more closely match the target distribution than those with CFG.

\subsection{Zero-init for ODE Solver}\label{sec:zero_init}
During the empirical validation of $s^\star$, we also studied how well the guided velocity from \ours matches the ground-truth. As shown in~\figref{fig:toy_jsd_norm}(b), during the early stages of training, the difference between both types of estimated velocities and the ground-truth velocity at $t=0$ is greater than when just using an all-zero velocity (do nothing), \ie,
\bea
\norm{\tilde{\vv}^\theta_0(x|y) - \vv_0^*(x|y)}_2^2 \geq \norm{\mathbf{0}-\vv_0^*(x|y)}_2^2.
\eea
Based on this observation, we propose zero-init, which zeros out the velocity for the first few steps of the ODESolver. As an estimation of all zeros for the velocity would be more accurate. Next, we investigate whether this behavior is specific to the small scale mixture of Gaussians dataset or could be generalized to real data. Specifically, we consider the experiment of ImageNet-256. 

\subsection{Validation beyond Mixture of Gaussians}

We experiment on the ImageNet-256~\cite{deng2009imagenet} benchmark where we train a class-conditioned flow model, \ie, given an input class label, generate an image from the class. We report results across multiple evaluation metrics, including \textit{Inception Score} (IS)~\cite{salimans2016improved}, \textit{Fréchet Inception Distance} (FID)~\cite{heusel2017gans}, \textit{sFID}~\cite{nguyen2022sliced}, \textit{Precision}, and \textit{Recall}. 

{\noindent\bf Validating Zero-Init.}
To study the impact of zero-init on a real dataset with flow matching models, we compare samples generated by the standard CFG with and without the zero-init throughout the training epochs, as shown in~\tabref{tab:training_steps}. Theoretically, if a model is well-trained, \textit{zero-init} could degrade its performance. For quick validation, we use a smaller version of DiT~\cite{peebles2023scalable} with 400M parameters as our base model and train it from scratch on the ImageNet-256 dataset using the Flow Matching loss~\cite{lipmanflow}.

The results show that CFG with \textit{Zero-Init} consistently outperforms standard CFG up to 160 epochs. This indicates that zeroing out the first step of the ODESolver is suitable during early training. Beyond 160 epochs, standard CFG surpasses Zero-Init, which is consistent with our hypothesis that as the velocity is trained to be more accurate, the benefit of zero-init lessens.

\begin{table}[t]
    \centering
    \resizebox{0.45\textwidth}{!}{
    \begin{tabular}{ll c c c c c}
        \toprule
        \multirow{2}{*}{Epochs} & \multirow{2}{*}{Methods} & \multicolumn{4}{c}{Metrics} \\
        \cmidrule(lr){3-7}
         &  & IS$\uparrow$ & 
         FID$\downarrow$ &
         sFID$\downarrow$ & Precision$\uparrow$ & Recall$\uparrow$ \\
        \midrule
        \multirow{2}{*}{10}
            & CFG & \textbf{53.27} & 28.57 & 18.52 & 0.61 & 0.36 \\
            & \cellcolor{gray!20}Zero-Init & \cellcolor{gray!20}52.78 & \cellcolor{gray!20}\textbf{28.55} & \cellcolor{gray!20}\textbf{17.32} & \cellcolor{gray!20}\textbf{0.62} & \cellcolor{gray!20}\textbf{0.37} \\
        \midrule
        \multirow{2}{*}{20}
            & CFG & \textbf{257.23} & 11.00 & 11.64 & 0.92 & 0.24 \\
            & \cellcolor{gray!20}Zero-Init & \cellcolor{gray!20}255.79 & \cellcolor{gray!20}\textbf{10.65} & \cellcolor{gray!20}\textbf{10.95} & \cellcolor{gray!20}\textbf{0.92} & \cellcolor{gray!20}\textbf{0.25} \\
        \midrule
        \multirow{2}{*}{40}
            & CFG & \textbf{339.39} & 12.61 & 11.17 & 0.94 & 0.23 \\
            & \cellcolor{gray!20}Zero-Init & \cellcolor{gray!20}338.40 & \cellcolor{gray!20}\textbf{12.29} & \cellcolor{gray!20}\textbf{10.47} & \cellcolor{gray!20}\textbf{0.94} & \cellcolor{gray!20}\textbf{0.24} \\
        \midrule
        \multirow{2}{*}{80}
            & CFG & 383.06 & 13.53 & 10.99 & 0.94 & 0.24 \\
            & \cellcolor{gray!20}Zero-Init & \cellcolor{gray!20}\textbf{383.45} & \cellcolor{gray!20}\textbf{12.18} & \cellcolor{gray!20}\textbf{10.39} & \cellcolor{gray!20}\textbf{0.94} & \cellcolor{gray!20}\textbf{0.26} \\
        \midrule
        \multirow{2}{*}{\textbf{160}}
            & CFG & \textbf{222.13} & \textbf{2.84} & \textbf{4.56} & \textbf{0.81} & \textbf{0.56} \\
            & \cellcolor{gray!20}Zero-Init & \cellcolor{gray!20}218.90 & \cellcolor{gray!20}2.85 & \cellcolor{gray!20}4.97 & \cellcolor{gray!20}0.80 & \cellcolor{gray!20}0.56 \\
        \bottomrule
    \end{tabular}}
    \caption{\textbf{Validation on ImageNet-256. 
    } We evaluate a model at different training stages and observe a turning point at 160 epochs, where zero-init results in poorer performance when the model converges. This experiment validates that high-dimensional models also suffer from inaccuracies in initial sampling.}
    \label{tab:training_steps}
\end{table}
\noindent\textbf{Validation of \ours.}  
To highlight the differences between our method and other classifier-free guidance approaches, we conduct experiments using a pre-trained SiT-XL model~\cite{ma2024sit} (700M parameters) on the ImageNet-256 benchmark. The model has not yet reached the "turning point" mentioned earlier. All experiments are performed using the standard guidance scale.

In~\tabref{tab:fid_is_comparison}, our results demonstrate that \ours achieves the best overall performance, outperforming both CFG++~\cite{chung2024cfgplusplus} and ADG~\cite{sadat2024eliminating} across key metrics. Specifically, \ours attains the highest \textit{Inception Score} of 258.87, highlighting its ability to generate diverse and high-quality images. Furthermore, \ours achieves the best \textit{FID Score} of 2.10 and \textit{sFID Score} of 4.59, indicating improved perceptual quality and stronger alignment with the target distribution. In terms of fidelity metrics, our method maintains a competitive \textit{Precision} of 0.80—comparable to both CFG and ADG—while achieving the highest \textit{Recall} of 0.61. This suggests that our approach better captures the underlying data distribution, leading to more representative and well-balanced samples. (Both ADG~\cite{sadat2024eliminating} and CFG++~\cite{chung2024cfgplusplus} are not designed for classifier-free guidance in Flow Matching, and therefore may not perform well.)

\section{Experiments}  
\label{Sec:exp}
In this section, we evaluate \ours on large-scale models for text-to-image (Lumina-Next~\cite{luminanext}, SD3~\cite{sd3}, SD3.5~\cite{sd3}, Flux~\cite{flux}) and text-to-video (Wan2.1~\cite{wan2.1} generation. Note: Flux is CFG-distilled, so directly applying classifier-free guidance may yield different results.


\begin{table}[t]
    \centering
    \resizebox{0.47\textwidth}{!}{
    \begin{tabular}{lccccc}
        \toprule
        Method & IS$\uparrow$ & FID$\downarrow$ & sFID$\downarrow$ & 
        Precision$\uparrow$ & Recall$\uparrow$ \\
        \midrule
        Baseline & 125.13 & 9.41 & 6.40 & 0.67 & 0.67\\
        w/ CFG & 257.03 & 2.23 & 4.61 & \textbf{0.81} & 0.59\\
        w/ ADG~\cite{sadat2024eliminating} & 257.92 & 2.37 & 5.51 & 0.80 & 0.58\\
        w/ CFG++~\cite{chung2024cfgplusplus} & 257.04 & 2.25 & 4.65 & 0.79 & 0.57\\
        \rowcolor{gray!20} \textbf{w/ \ours} & \textbf{258.87} & \textbf{2.10} & \textbf{4.59} & 0.80 & \textbf{0.61}\\
        \bottomrule
    \end{tabular}}
    \caption{\textbf{Comparison of different guidance strategy on ImageNet-256 benchmark.} Lower FID is better ($\downarrow$) and higher IS is better ($\uparrow$). Baseline here denotes using the conditional prediction only.}
    \vspace{-2mm}
    \label{tab:fid_is_comparison}
\end{table}
\begin{table}[t]
\centering
\resizebox{0.45\textwidth}{!}{
\begin{tabular}{l c c c}
\toprule
Model & Method & Aesthetic Score$\uparrow$ & Clip Score$\uparrow$ \\
\midrule
\multirow{2}{*}{Lumina-Next~\cite{luminanext}} & CFG & 6.85 & 34.09 \\
                          & \cellcolor{gray!20} \ours & \cellcolor{gray!20} \textbf{7.03} & \cellcolor{gray!20} \textbf{34.37} \\
\midrule
\multirow{2}{*}{SD3~\cite{sd3}} & CFG & 6.73 & 34.00 \\
                          & \cellcolor{gray!20} \ours & \cellcolor{gray!20} \textbf{6.80} & \cellcolor{gray!20} \textbf{34.11} \\
\midrule
\multirow{2}{*}{SD3.5~\cite{sd3}} & CFG & 6.96 & 34.60 \\
                          & \cellcolor{gray!20} \ours & \cellcolor{gray!20} \textbf{7.10} & \cellcolor{gray!20} \textbf{34.68} \\
\midrule
\multirow{2}{*}{Flux~\cite{flux}} & CFG & 7.06 & 34.60 \\
                          & \cellcolor{gray!20} \ours & \cellcolor{gray!20} \textbf{7.12} & \cellcolor{gray!20} \textbf{34.69} \\
\bottomrule
\end{tabular}}
\caption{\textbf{Quantitative evaluation of Text-to-Image generation}, using Lumina-Next, Stable Diffusion 3, Stable Diffusion 3.5, and Flux. The evaluation is based on \textit{Aesthetic Score} and \textit{CLIP Score} as key metrics. Results indicate that \ours consistently enhances image quality and improves alignment with textual prompts across different models.}
\label{tab:t2i}
\vspace{-0.2cm}
\end{table}
\begin{figure*}[t]
\centering
\includegraphics[width=0.99\linewidth]{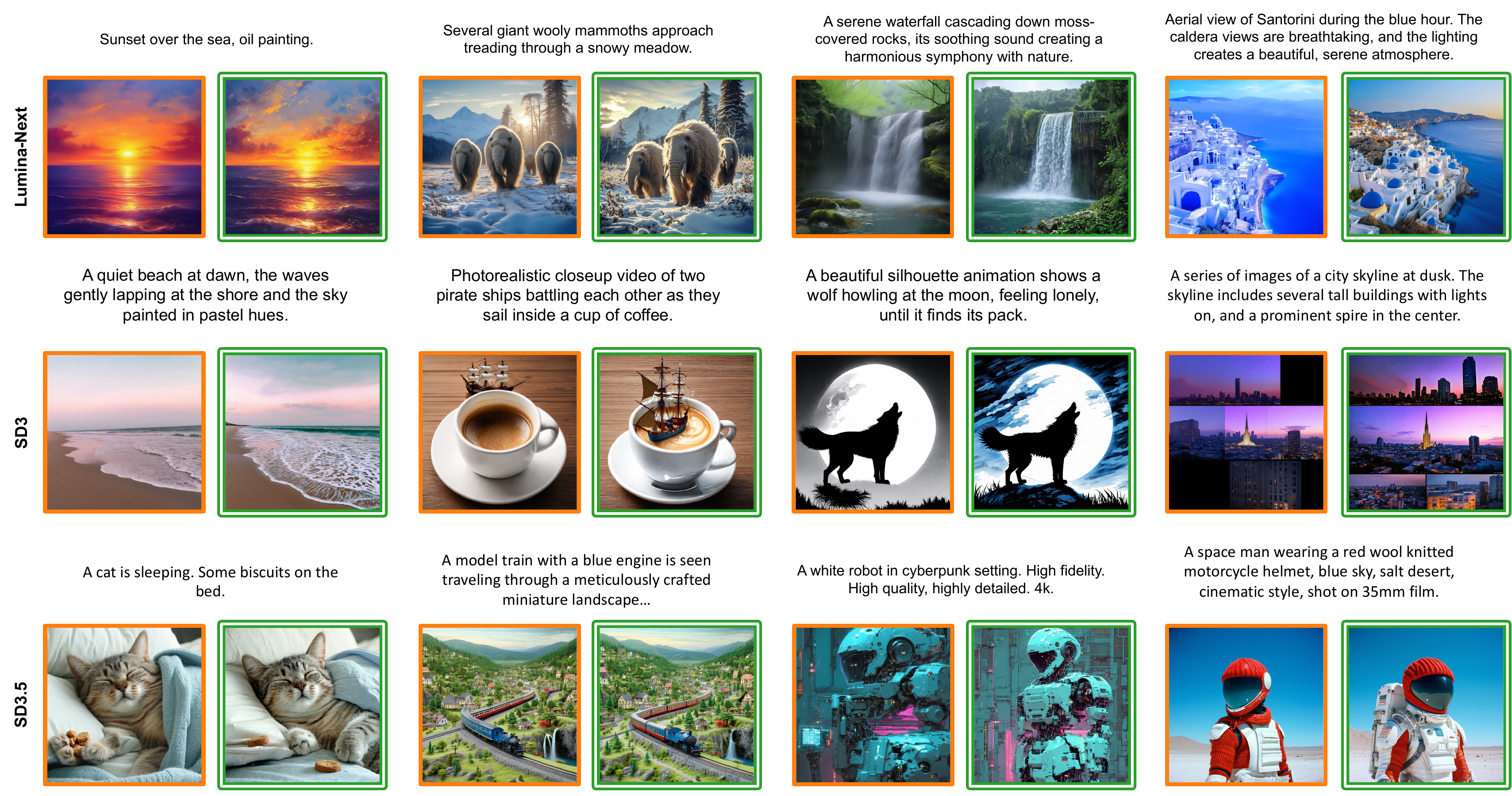}
\caption{\textbf{Qualitative comparisons between CFG and \ours.} Experiments are conducted using Lumina-Next, Stable Diffusion 3, and Stable Diffusion 3.5, with each model evaluated under its recommended optimal sampling steps and guidance scale settings. CFG results are shown in \textcolor{orange}{orange} and 
{\bf Ours} are highlighted in \textcolor{ForestGreen}{green} boxes.
}
\vspace{-3mm}
\label{fig:t2i_compare}
\end{figure*}

\subsection{Text-to-Image Generation}
To evaluate the effectiveness of our proposed method, \ours, in continuous class-conditional image generation, we conducted experiments using four state-of-the-art flow matching models: Lumina-Next~\cite{luminanext}, SD3~\cite{sd3}, SD3.5~\cite{sd3}, and Flux~\cite{flux}. These models were selected for their strong performance in class-conditional image synthesis. We applied both \ours and the standard CFG under default settings on a diverse set of self-curated prompts, allowing us to provide fair comparisons.\\[1ex]
\noindent\textbf{Quantitative Evaluation.} 
\tabref{tab:t2i} presents the quantitative comparison of \ours and CFG across all tested models. The results indicate that \ours consistently achieves superior performance, as evidenced by higher \textit{Aesthetic Score}~\cite{talebi2018nima} and \textit{CLIP Score}~\cite{radford2021learning,peebles2023image}. The improvement in \textit{Aesthetic Score} suggests that \ours enhances the visual appeal of generated images, producing outputs with more coherent textures, lighting, and structure. Additionally, the increase in \textit{CLIP Score} demonstrates that \ours improves text-image alignment, ensuring that generated images better capture the semantics of the given prompts. These results validate the effectiveness of our proposed modifications in refining the quality of diffusion-based generation.\\ 

\noindent\textbf{Qualitative Evaluation.}~\figref{fig:t2i_compare} provides a comparisons of the images generated using \ours and vanilla CFG. Our method produces high-fidelity outputs that exhibit richer details, sharper textures, and better preservation of object structures compared to the baseline CFG. Notably, \ours mitigates common artifacts observed in CFG-generated images, particularly those that introduce unintended distortions or elements unrelated to the given prompt. This reduction in artifacts highlights the robustness of \ours in preserving semantic consistency, ensuring that generated images adhere more closely to the given prompts. 
\begin{table}[t]
    \centering
    \resizebox{0.45\textwidth}{!}{
    \begin{tabular}{lcccc}
        \toprule
        Method & Color$\uparrow$ & Shape$\uparrow$ & Texture$\uparrow$ & 
        Spatial$\uparrow$ \\
        \midrule
        Lumina-Next~\cite{luminanext} & 0.51 & 0.34 & 0.41 & 0.19 \\
        \rowcolor{gray!20} \textbf{+ \ours} & \textbf{0.52} & \textbf{0.36} & \textbf{0.45} & \textbf{0.29} \\
        \midrule
        SD3~\cite{sd3} & 0.81 & 0.57 & 0.71 & \textbf{0.31} \\
        \rowcolor{gray!20} \textbf{+ \ours} & \textbf{0.83} & \textbf{0.58} & \textbf{0.72} & \textbf{0.31} \\
        \midrule
        SD3.5~\cite{sd3} & 0.76 & 0.59 & 0.70 & 0.27\\
        \rowcolor{gray!20} \textbf{+ \ours} & \textbf{0.78} & \textbf{0.60} & \textbf{0.71} & \textbf{0.28}\\
        \bottomrule
    \end{tabular}}
    \caption{\textbf{Quantitative evaluation on T2I-CompBench~\cite{huang2025t2icompbench++}}, using Lumina-Next, Stable Diffusion 3, and Stable Diffusion 3.5. Compared to CFG, \ours demonstrates consistent improvements across all evaluated aspects. }
    \vspace{-5mm}
    \label{tab:t2i_comp_bench}
\end{table}

Additionally, we observe that \ours have better color consistency and level of detail, reducing blurry artifacts that are sometimes present in CFG-based outputs. These improvements are particularly evident in complex prompts that require the precise rendering of intricate textures or fine-grained semantic attributes. Further visual comparisons and additional generated samples can be found in the Appendix. \\[1ex]
\noindent\textbf{Benchmark Results.}  
We compare our method against standard CFG across three different Flow Matching models using T2I-CompBench~\cite{huang2023t2icompbench,huang2025t2icompbench++}. As shown in \tabref{tab:t2i_comp_bench}, integrating \ours leads to notable improvements in \textit{Color}, \textit{Shape}, and \textit{Texture} quality in generated images. Meanwhile, the \textit{Spatial} dimension remains comparable to the baseline, indicating that \ours improves image fidelity without compromising structural coherence.\\[1ex]
\noindent\textbf{User Study.} To further assess \ours, we conduct a user study to compare its performance against standard CFG across various flow matching models. Participants were presented with image pairs generated using both \ours and CFG and were asked to evaluate them based on three key aspects: \textit{detail preservation}, \textit{color consistency}, and \textit{image-text alignment}. The overall preference score was then computed as the percentage of times \ours was favored over CFG.
\begin{figure}[t]
\centering
\includegraphics[width=0.99\linewidth]{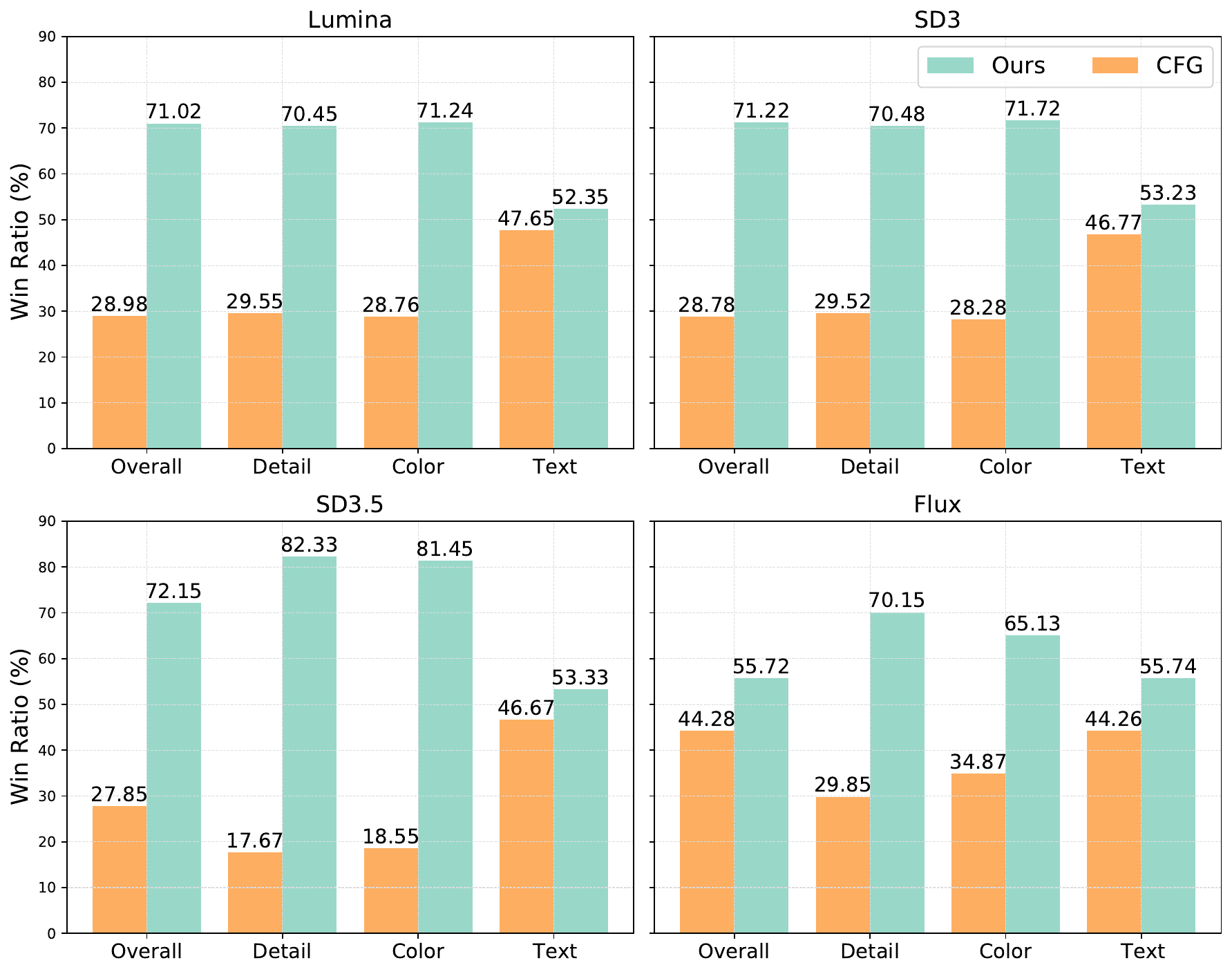}
\caption{\textbf{User study on Lumina-Next, Stable Diffusion 3, Stable Diffusion 3.5, and Flux}. The win rate of our method compared to CFG is presented.}
\label{fig:user_study}
\vspace{-3mm}
\end{figure}

\begin{table*}[t]
    \centering
    \resizebox{0.99\textwidth}{!}{
    \begin{tabular}{lccccccccccccccccc}
        \toprule
        Method & 
        Total Score & subject consistency  & 
        aesthetic quality & 
        imaging quality &  
        color & 
        spatial relationship & 
        temporal style & 
        motion smoothness & 
 \\
        \midrule
        Vchitect-2.0 [2B]~\cite{fan2025vchitect} & 81.57 & 61.47 &  65.60 & 86.87 & 86.87 & 54.64 &  25.56 & 97.76 \\
        CogVideoX-1.5 [5B]~\cite{yang2024cogvideox} & 82.17 & 96.87 & 62.79 & 65.02 & 87.55 & 80.25 &  25.19 & 98.31\\
        \midrule
        Wan2.1 [14B]~\cite{wan2.1}& 83.99 & 93.33  & 69.13  & 67.48 & 83.43 & \textbf{80.46} & 25.90 & \textbf{98.05}\\
        \rowcolor{gray!20} \textbf{w/ \ours} & \textbf{84.06} & \textbf{93.34} & \textbf{69.22} & \textbf{67.55} & \textbf{85.39} & 79.28 & \textbf{25.98} & 98.00\\
        \midrule
        Wan2.1 [1B]~\cite{wan2.1} & 80.52 & 93.89 &  61.67 & 65.40 & 87.57 & 72.75 &  \textbf{24.13} & 97.24 \\
        \rowcolor{gray!20} \textbf{w/ \ours} & \textbf{80.91} & \textbf{94.93} & \textbf{64.24} & \textbf{68.13} & \textbf{89.36} & \textbf{73.84} & 23.36 & \textbf{98.16}\\
        \bottomrule
    \end{tabular}}
    \vspace{-2mm}
    \caption{\textbf{Qualitative evaluation on VBench~\cite{huang2023vbench}.}  
We use the Wan-2.1~\cite{wan2.1} model as our base model. Compared to vanilla CFG, \ours improves both frame quality and overall video smoothness. 
 }
 \vspace{-2mm}
    \label{tab:vbench}
\end{table*}
\begin{figure*}[t]
\centering
\includegraphics[width=0.99\linewidth]{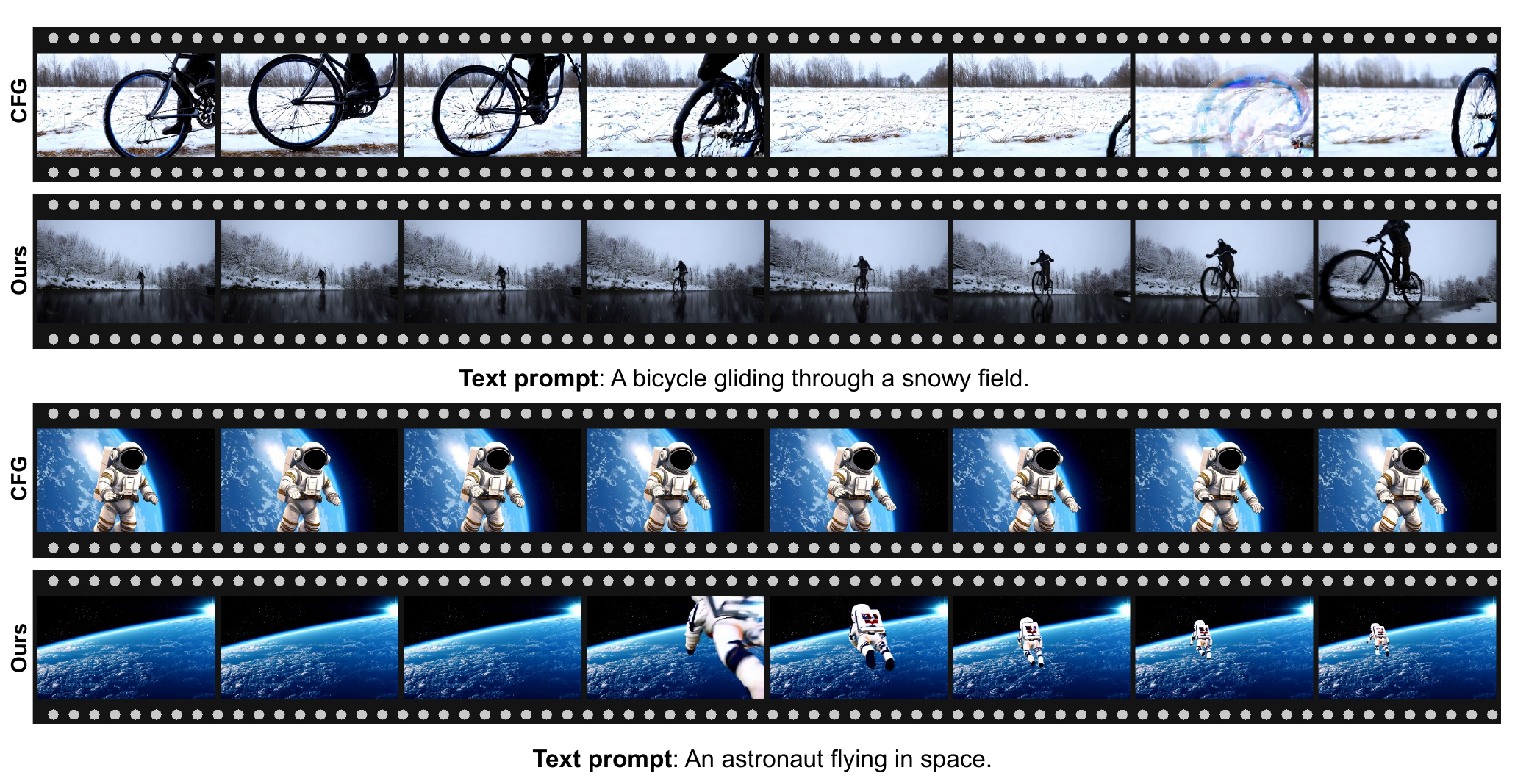}
\caption{\textbf{Qualitative comparisons between \ours and CFG.} Experiments are conducted using Wan-2.1 [1B]~\cite{wan2.1}, under its recommended optimal sampling steps and guidance scale settings.
}
\vspace{-3mm}
\label{fig:video_qualitative}
\end{figure*}

\figref{fig:user_study} summarizes the results. \ours demonstrates a clear advantage over CFG across all tested models. Notably, our method achieves the highest win ratio on SD3.5, with an overall win ratio score of \textbf{72.15\%}, primarily driven by significant improvements in \textit{detail preservation} (\textbf{82.33\%}) and \textit{color fidelity} (\textbf{81.45\%}). This suggests that \ours effectively enhances fine-grained structures and maintains consistent color distributions compared to its baseline. In terms of \textit{image-text alignment}, \ours also performs favorably, surpassing CFG in most cases. Specifically, with CFG++, SD3.5 exhibits strong text-image consistency (\textbf{53.33\%} win ratio), indicating that our method improves coherence between generated images and textual prompts across different architectures.

\subsection{Text-to-Video Generation}
To further evaluate the effectiveness of our proposed method, we further conduct experiments on the text-to-video generation task, using the most recent state-of-the-art model, Wan-2.1~\cite{wan2.1}.

\noindent\textbf{Benchmark Results.}  
As shown in \tabref{tab:vbench}, we evaluate our method using all metrics from VBench~\cite{huang2023vbench,huang2024vbench++}. Compared to CFG, Wan-2.1 [1.3B] equipped with \ours achieves a higher \textit{Total Score}. Specifically, our method improves \textit{Aesthetic Quality} by \textbf{2.57} and \textit{Imaging Quality} by \textbf{2.73}, indicating that \ours enhances video fidelity. Additionally, \ours improves \textit{Motion Smoothness} (\textbf{+0.92}) and \textit{Spatial Relationship} (\textbf{+1.09}), demonstrating superior temporal coherence and spatial understanding.  
However, we also observe a decrease in \textit{Temporal Style} (\textbf{-0.77}), which can be attributed to the base model's poor capability in generating stylized videos.\\[1ex]
\noindent\textbf{Qualitative Evaluation.}  
\figref{fig:video_qualitative} presents a visual comparison between \ours and CFG on Wan2.1-14B~\cite{wan2.1}, demonstrating that the refined velocity produced by our method leads to more plausible content with natural motion.

\subsection{Ablation Studies}
\begin{figure}[t]
    \centering
    \setlength{\tabcolsep}{0pt}
    \begin{tabular}{cc}
    \includegraphics[width=0.5\linewidth, height=3.19cm]{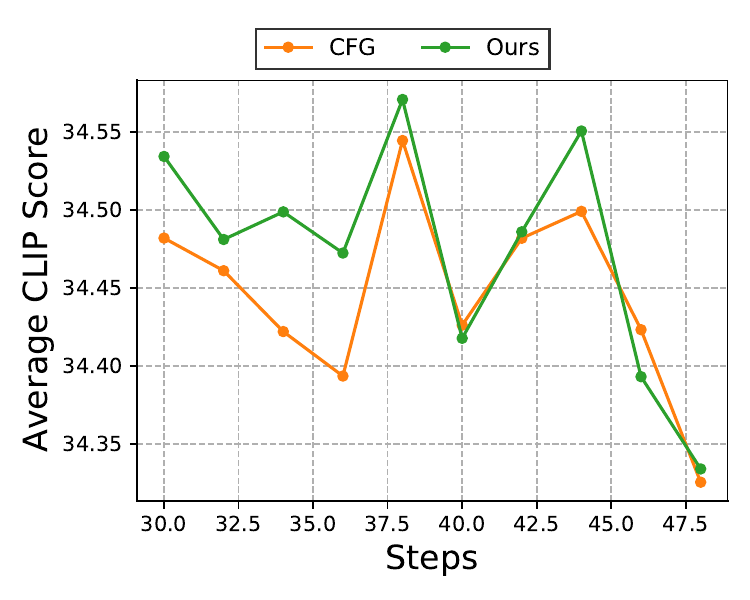} &
    \includegraphics[width=0.5\linewidth, height=3.19cm]{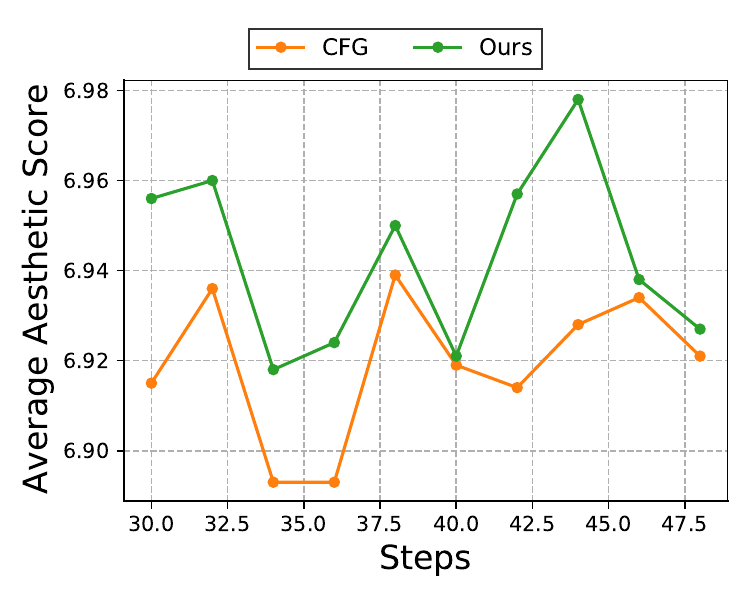}\\
    (a) & (b)
    \end{tabular}
    \vspace{-2mm}
    \caption{\textbf{Abalation study on different sampling steps.} Comparison of \textit{CLIP Score} and \textit{Aesthetic Score} between our method and CFG across different sampling steps.
    }
    \vspace{-2mm}
    \label{fig:ablation_steps}
\end{figure}

\begin{figure}[t]
    \centering
    \setlength{\tabcolsep}{0pt}
    \begin{tabular}{cc}
    \includegraphics[width=0.5\linewidth, height=3.19cm]{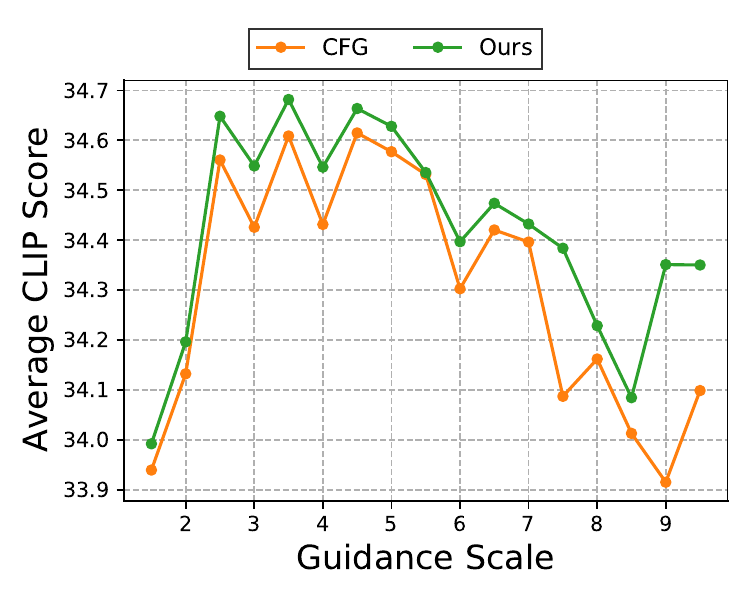} &
    \includegraphics[width=0.5\linewidth, height=3.19cm]{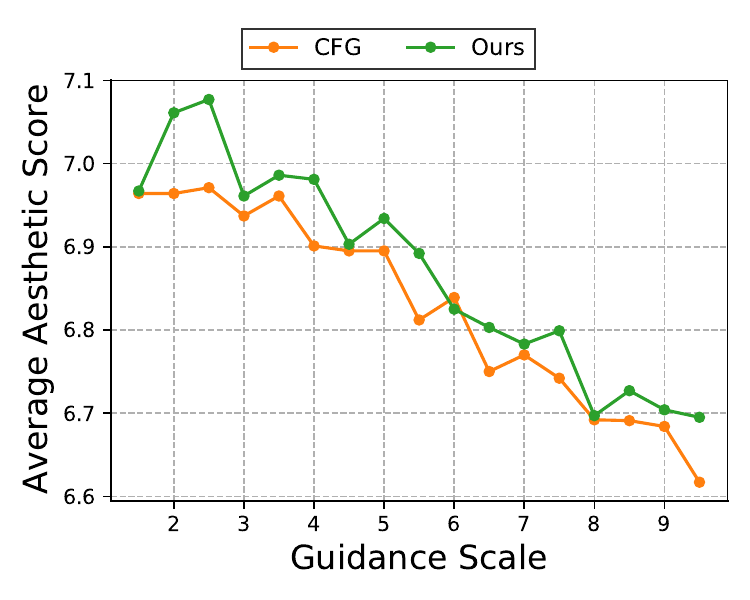}\\
    (a) & (b)
    \end{tabular}
    \vspace{-2mm}
    \caption{\textbf{Abalation study on different guidance scale.} Comparison of \textit{CLIP Score} and \textit{Aesthetic Score} between our method and CFG across different guidance scale.
    }
    \vspace{-4mm}
    \label{fig:ablation_scale}
\end{figure}

\noindent\textbf{Different Sampling Steps.}
To further investigate the impact of sampling steps, we conduct experiments using a strong baseline, SD3.5. As shown in ~\figref{fig:ablation_steps}, our method consistently enhances both image quality and text-matching accuracy across different sampling steps, demonstrating its robustness in varied sampling settings.\\[1ex]
\noindent\textbf{Different Guidance Scale.}
As shown in ~\figref{fig:ablation_scale}, we conduct an experiment using SD3.5 to analyze the impact of different guidance scales on our method. The results in sub-figure (a) demonstrate that \ours achieves higher CLIP scores than CFG across all guidance scales, validating its effectiveness in improving image-text alignment. Similarly, sub-figure (b) shows that our method consistently attains higher aesthetic scores, highlighting its ability to generate visually appealing images.\\[1ex]


\noindent\textbf{Effectiveness of \ours.}  
We conduct an ablation study using SD3.5 as the baseline to assess the impact of each component in \ours. As shown in \tabref{tab:effectiveness}, we compare four variants: vanilla CFG, CFG with zero-init, pure optimized scaler, and \ours. The optimized scaler reduces the gap between predicted and true velocity, enhancing stability, while zero-init improves performance by skipping the first step. Combining both, \ours achieves the highest Aesthetic Score (7.10) and CLIP Score (34.68), outperforming all variants. These results confirm that both modifications help to enhance image quality and text alignment.
\\[1ex]
\noindent\textbf{Effect of Zeroing Out Initial Steps.}  To assess whether extending the zero-out strategy beyond the first step can further improve performance, we conduct an ablation study using Lumina-Next, SD3, and SD3.5, as shown in \tabref{tab:zero_out_ablate}. The results indicate that for some models, zeroing out can be beneficial not only in the first step but also in the initial few steps. Specifically, both Lumina-Next and SD3 achieve optimal performance when the first 2 steps are zeroed out. However, SD3.5 exhibits a decline in performance when a higher proportion of initial steps are zeroed out, suggesting that SD3.5 is well-trained and approaching convergence.\\[1ex]
\noindent\textbf{Computational Cost.}~\tabref{tab:FLOPs} summarizes the FLOPs and GPU memory usage of our method. For 720p and 480p resolutions using Wan2.1, our method requires 19.4M and 0.84M FLOPs, with memory usage of 18.46MB and 8.00MB. For SD3, the computational cost is significantly lower at 1024×1024 and 512×512, requiring only 1.6e$^{-3}$M and 4.1e$^{-4}$M FLOPs, with memory usage of 64KB and 16KB. These results confirm that \ours introduces minimal computational costs.

\begin{table}[t]
    \centering
    \resizebox{\linewidth}{!}{
    \begin{tabular}{lcccc}
        \toprule
        Metrics & w/ CFG & w/ CFG-Zero & w/ Scaler & w/ \ours \\
        \midrule
        Aesthetic Score & 6.96 & \underline{7.00} & 6.96 & \textbf{7.10}\\
        CLIP Score & 34.60 & \underline{34.65} & 34.64 & \textbf{34.68}\\
        \bottomrule
    \end{tabular}}
    \vspace{-1mm}
    \caption{\textbf{Effectiveness of \ours}. Comparison of vanilla CFG, CFG with zero-init, dynamic scaling, and \ours, highlighting the impact of \textit{zero-init} and \textit{dynamic scaling} in improving performance.}
    \label{tab:effectiveness}
\end{table}
\begin{table}[t]
\centering
\resizebox{\linewidth}{!}{
\begin{tabular}{l c c c}
\toprule
Model & Zero-out / Total (steps) & Aesthetic Score$\uparrow$ & Clip Score$\uparrow$ \\
\midrule
\multirow{3}{*}{Lumina-Next~\cite{luminanext}} & First 3 / 30 & 6.78 & 32.86 \\
& First 2 / 30 & \textbf{7.06} & \textbf{34.73} \\
                          &  First 1 / 30 &  7.03 &  34.37 \\
\midrule
\multirow{3}{*}{SD3~\cite{sd3}} & First 3 / 28 & 6.95 & 34.01 \\
& First 2 / 28 & \textbf{6.98} & \textbf{34.33} \\
                          & First 1 / 28 &  6.80 &  34.11 \\
\midrule
\multirow{3}{*}{SD3.5~\cite{sd3}}
& First 3 / 28 & 6.78 & 34.02 \\
& First 2 / 28 & 6.99 & 34.54 \\
                          & First 1 / 28 &  \textbf{7.10} & \textbf{34.68} \\
\bottomrule
\end{tabular}}
\vspace{-1mm}
\caption{\textbf{Ablation study on zero-out steps.} For SD3.5~\cite{sd3}, more initial zero-out steps lead to worse performance, while Lumina-Next~\cite{luminanext} and SD3~\cite{sd3} achieve the highest ~\textit{Aesthetic Score} and ~\textit{Clip Score} with first 7\% zero out. }
\label{tab:zero_out_ablate}
\end{table}
\begin{table}[t]
    \centering
    \resizebox{\linewidth}{!}{
    \begin{tabular}{lcccc}
        \toprule
        Resolution & 81x1280x720 & 81x832x480 & 1024x1024 & 512x512 \\
        \midrule
        FLOPs & 19.4 M & 0.84 M & 1.6$e^{-3}$ M & 4.1$e^{-4}$ M\\
        Memory & 18.46 MB & 8.00 MB & 64 KB & 16 KB\\
        \bottomrule
    \end{tabular}}
    \vspace{-1mm}
    \caption{\textbf{Computational costs.} FLOPs~\cite{jain2020benchmarking} and GPU memory usage of our method for 5-second video generation at 720p/480p using Wan2.1~\cite{wan2.1}, and at 1024/512 resolution using SD3~\cite{sd3}.
    \vspace{-.88mm}
}
    \label{tab:FLOPs}
\end{table}

\section{Conclusion}
We introduce \ours, an improved classifier-free guidance method for flow-matching diffusion models. \ours addresses CFG’s limitations with two key techniques: (1) an optimized scale factor for accurate velocity estimation and (2) a zero-init technique to stabilize early sampling. Theoretical analysis and extensive experiments on text-to-image (SD3.5, Lumina-Next, Flux) and text-to-video (Wan-2.1) models show that \ours outperforms standard CFG, achieving higher aesthetic scores, better text alignment, and fewer artifacts. Ablation studies further validate the effectiveness of both components in improving sample quality without significant computational cost.

{
    \small
    \bibliographystyle{ieeenat_fullname}
    \bibliography{ref}
}

\clearpage
\onecolumn 

\section*{Appendix}

\setcounter{section}{0}
\renewcommand{\theHsection}{A\arabic{section}}
\renewcommand{\thesection}{A\arabic{section}}
\renewcommand{\thetable}{A\arabic{table}}
\setcounter{table}{0}
\setcounter{figure}{0}
\renewcommand{\thetable}{A\arabic{table}}
\renewcommand\thefigure{A\arabic{figure}}
\renewcommand{\theHtable}{A.Tab.\arabic{table}}
\renewcommand{\theHfigure}{A.Abb.\arabic{figure}}
\renewcommand\theequation{A\arabic{equation}}
\renewcommand{\theHequation}{A.Abb.\arabic{equation}}


\section{Additional Experiments}
\subsection{Experiments on Mixed Gaussian}
\begin{figure*}[ht]
    \centering
    \setlength{\tabcolsep}{0pt} 
    \renewcommand{\arraystretch}{0}
    \begin{tabular}{@{}ccccccc@{}}
    \multirow{1}{*}[1.85cm]{\rotatebox[origin=c]{90}{CFG}} &
        \includegraphics[width=0.16\linewidth]{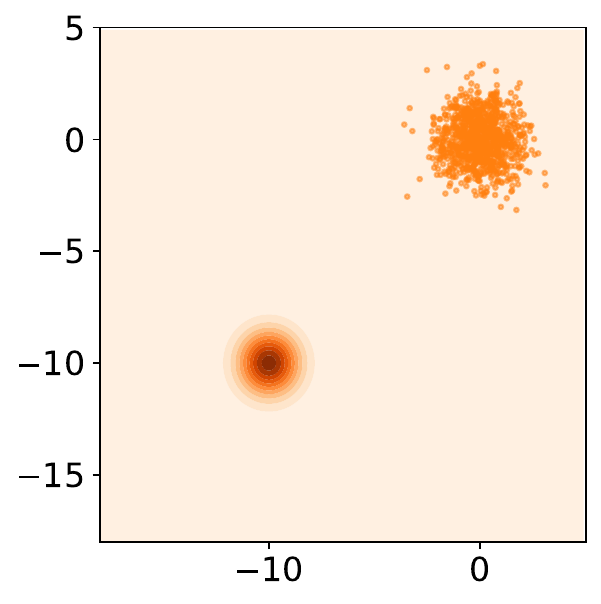} & 
        \includegraphics[width=0.16\linewidth]{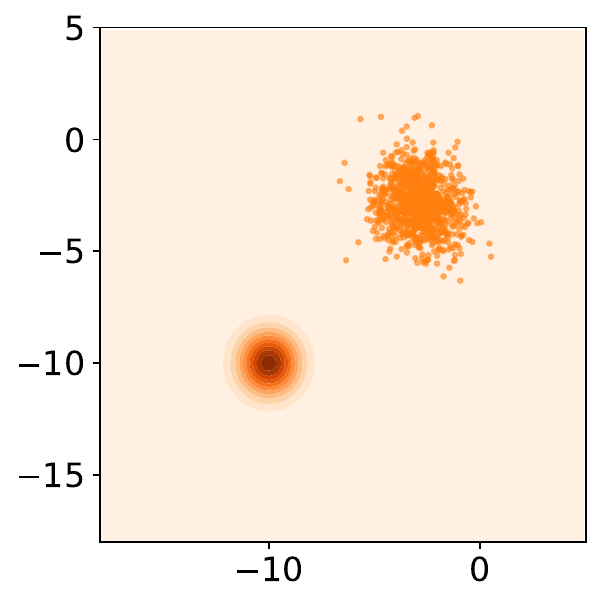} & 
        \includegraphics[width=0.16\linewidth]{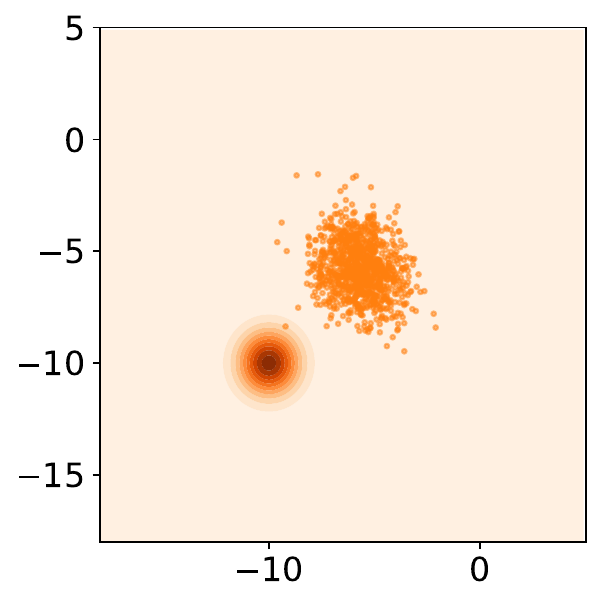} & 
        \includegraphics[width=0.16\linewidth]{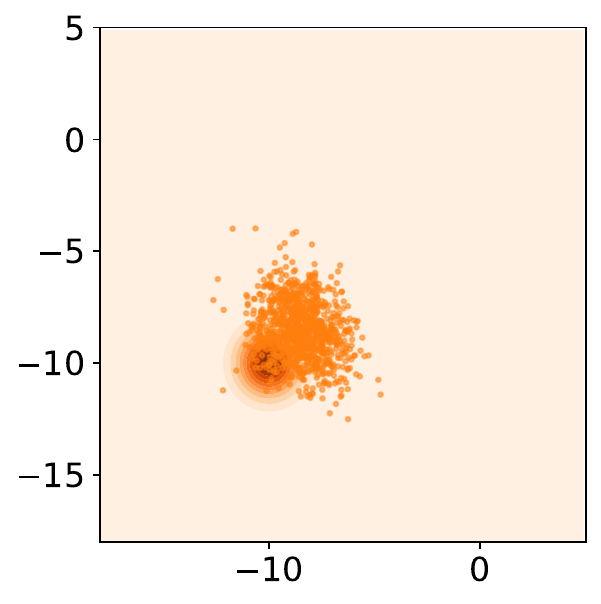} & 
        \includegraphics[width=0.16\linewidth]{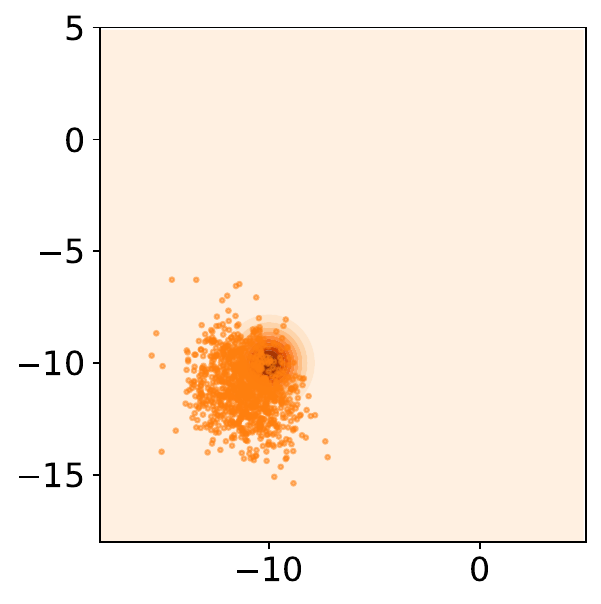} & 
        \includegraphics[width=0.16\linewidth]{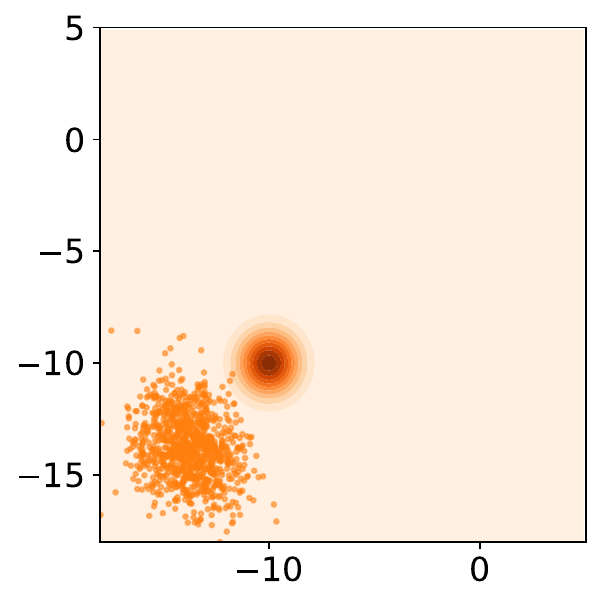} \\
    \multirow{1}{*}[2.0cm]{\rotatebox[origin=c]{90}{Cond}} &
        \includegraphics[width=0.16\linewidth]{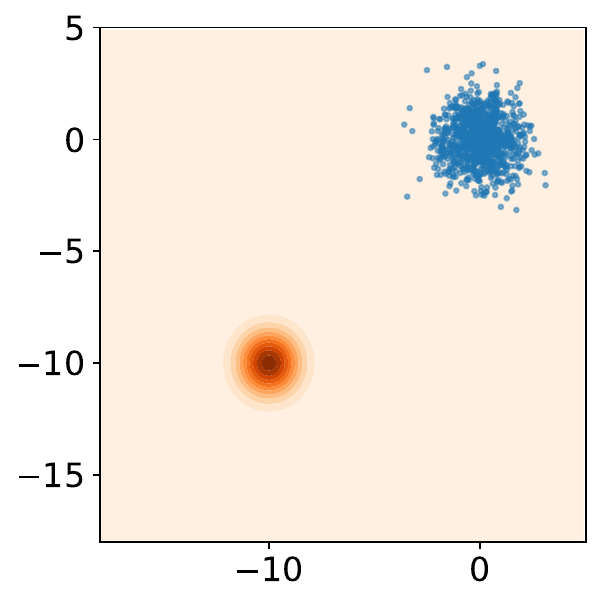} & 
        \includegraphics[width=0.16\linewidth]{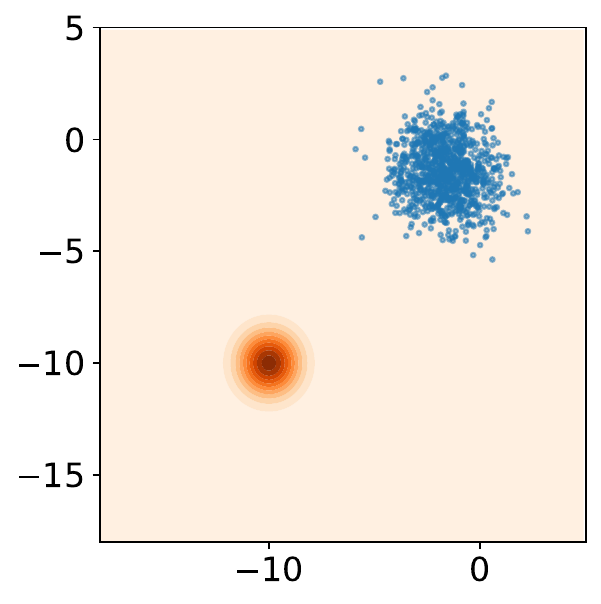} & 
        \includegraphics[width=0.16\linewidth]{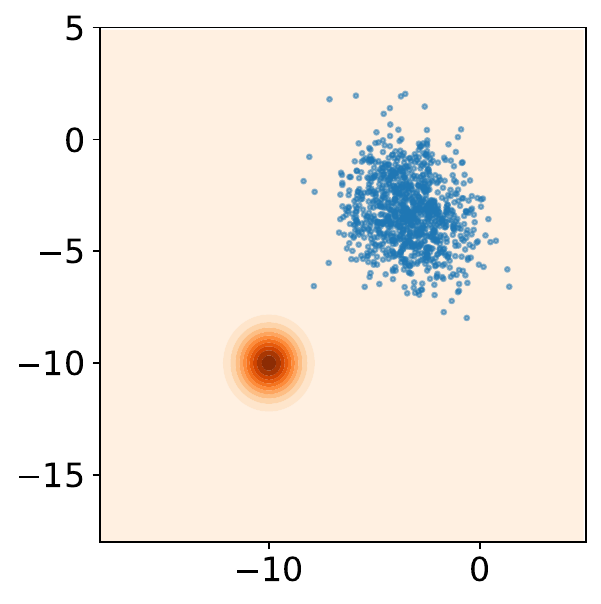} & 
        \includegraphics[width=0.16\linewidth]{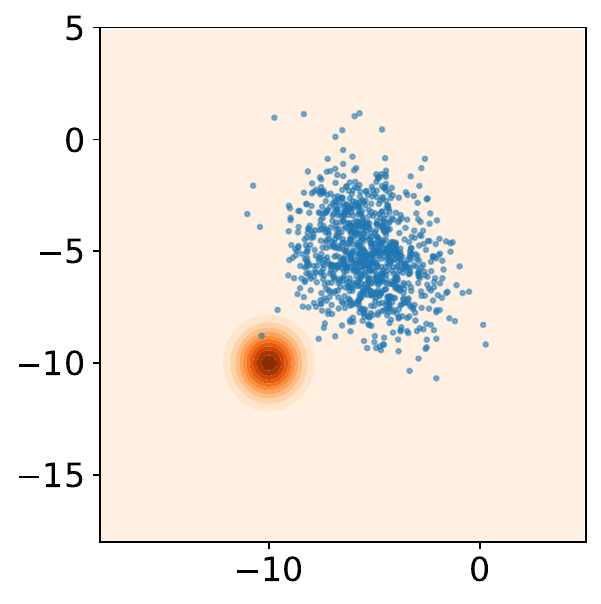} & 
        \includegraphics[width=0.16\linewidth]{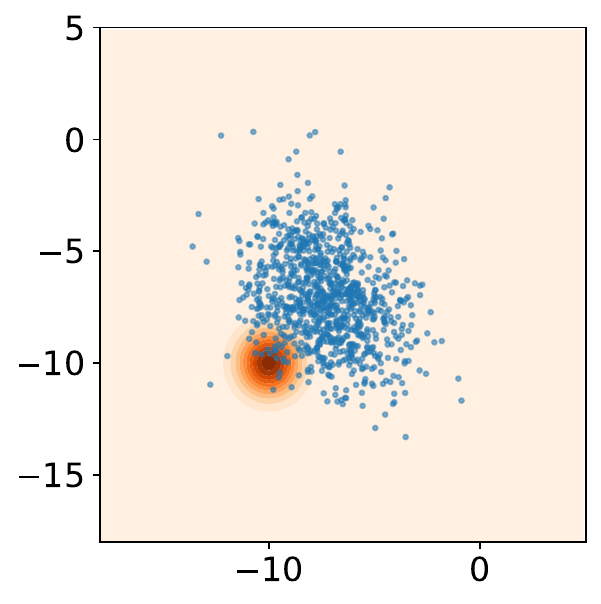} & 
        \includegraphics[width=0.16\linewidth]{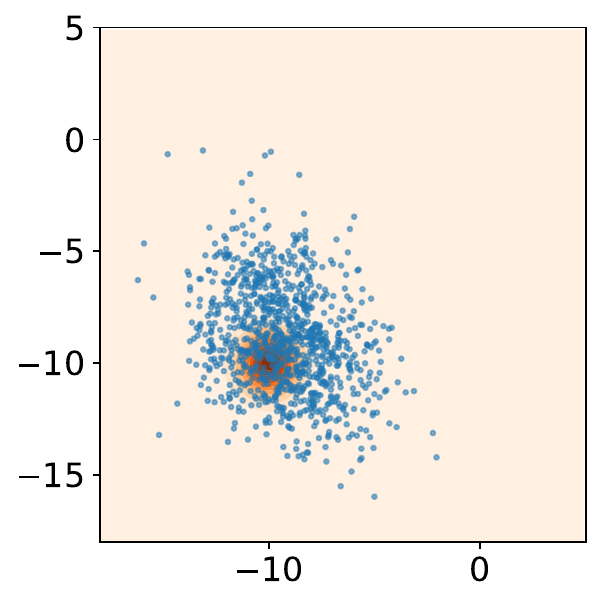} \\
    \multirow{1}{*}[2.4cm]{\rotatebox[origin=c]{90}{\ours}} &
        \includegraphics[width=0.16\linewidth]{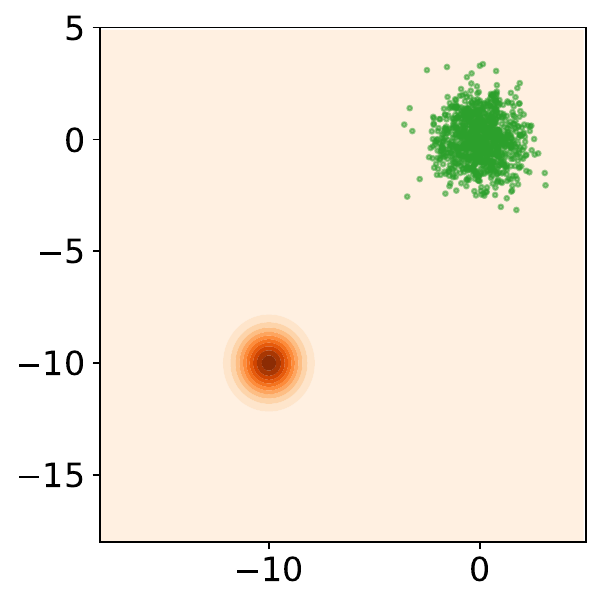} & 
        \includegraphics[width=0.16\linewidth]{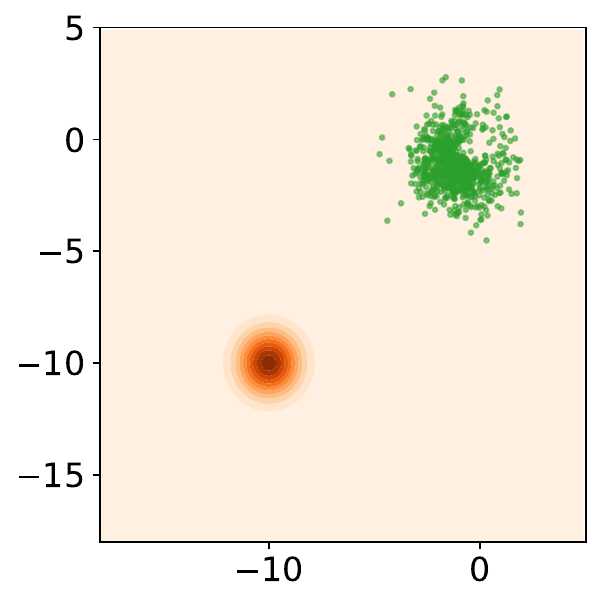} & 
        \includegraphics[width=0.16\linewidth]{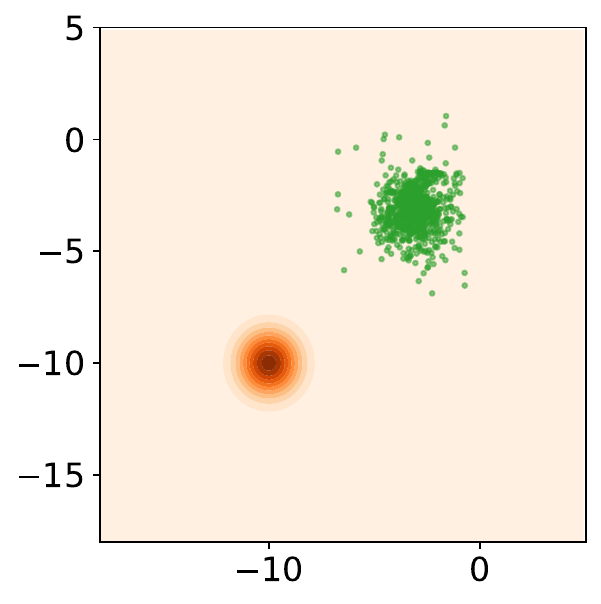} & 
        \includegraphics[width=0.16\linewidth]{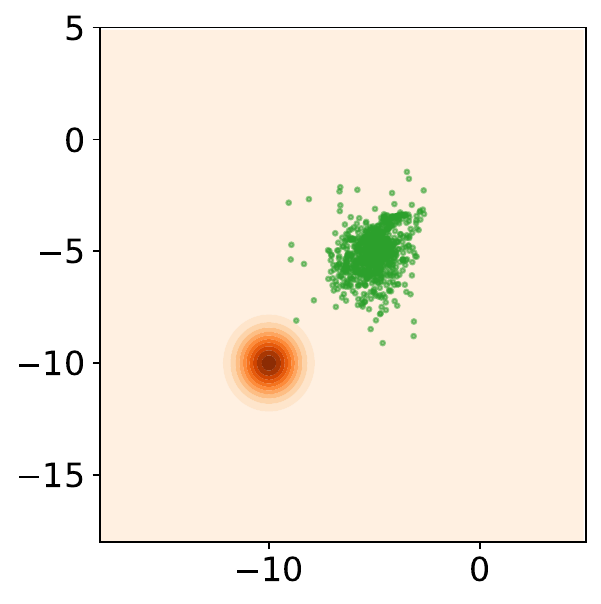} & 
        \includegraphics[width=0.16\linewidth]{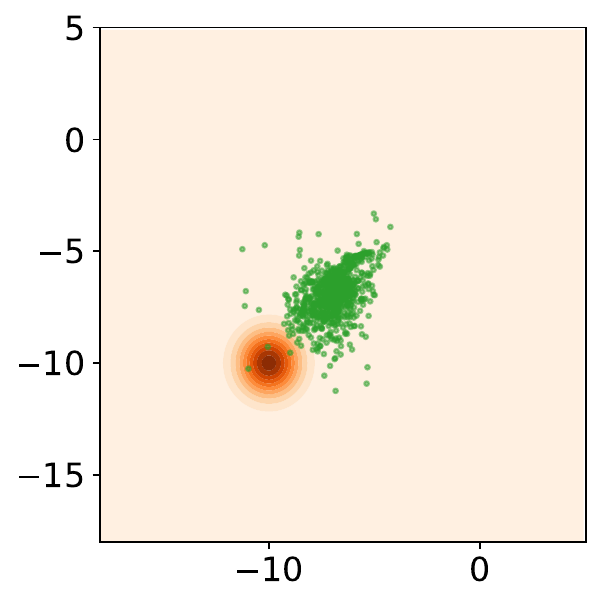} & 
        \includegraphics[width=0.16\linewidth]{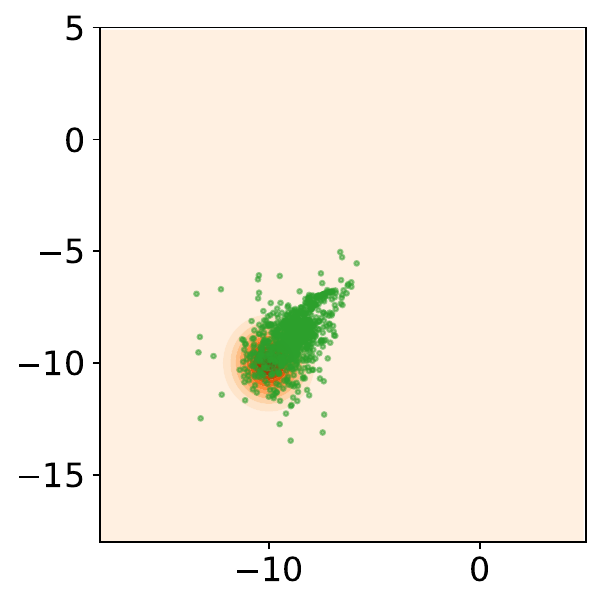} \\
        & (a) $t=0$ & (b) $t=0.2$ & (c) $t=0.4$ & (d) $t=0.6$ & (e) $t=0.8$ & (f) $t=1.0$
    \end{tabular}
    \caption{\textbf{Flow sampling trajectory.} Each panel shows the sample trajectories at a different time step.}
    \label{fig:flow_trajectory}
\end{figure*}

\noindent\textbf{Comparison of flow trajectory.} We present the flow sampling trajectories with 10 steps of different methods in~\figref{fig:flow_trajectory}. As shown in the last column, samples guided by CFG move across the target distribution, while using only Cond leads to high variance in the sampled distribution. In contrast, \ours effectively guides the samples toward the target distribution without excessive variance.

\begin{figure*}[ht]
    \centering
    \setlength{\tabcolsep}{0pt}
    \begin{tabular}{cccc}
    \includegraphics[width=0.25\linewidth,]{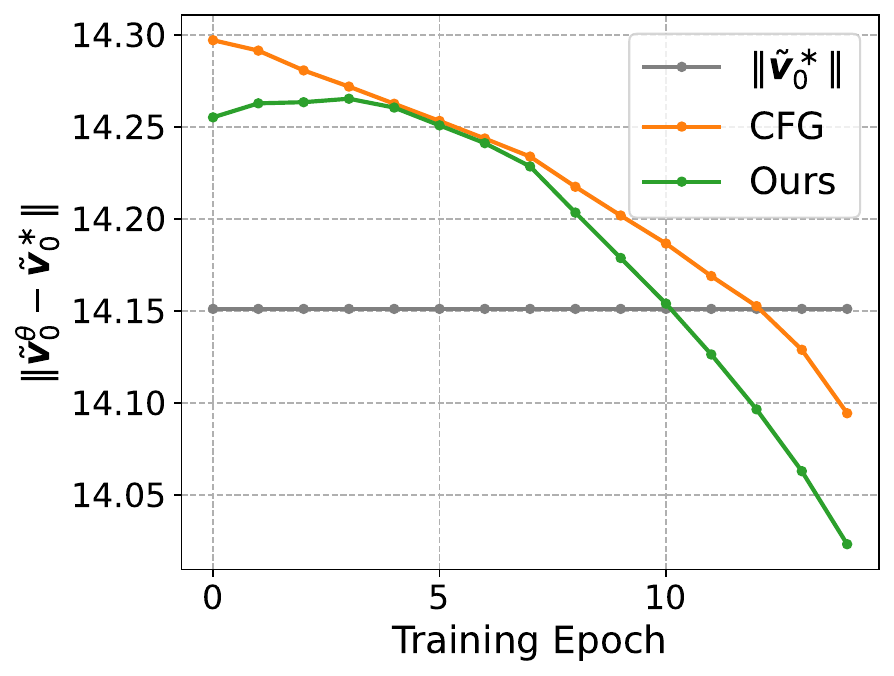} &
    \includegraphics[width=0.25\linewidth,]{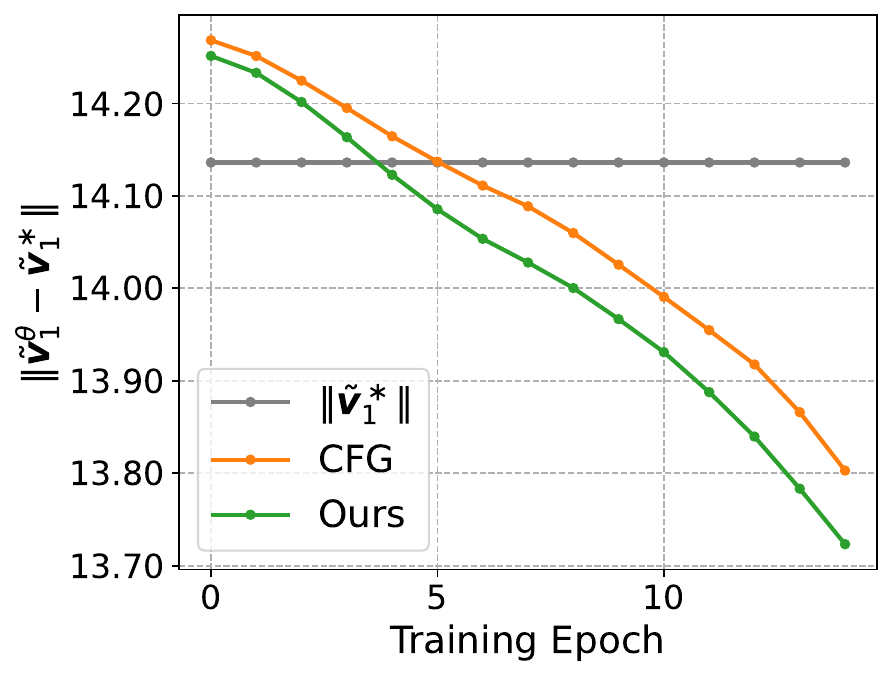} &
    \includegraphics[width=0.25\linewidth,]{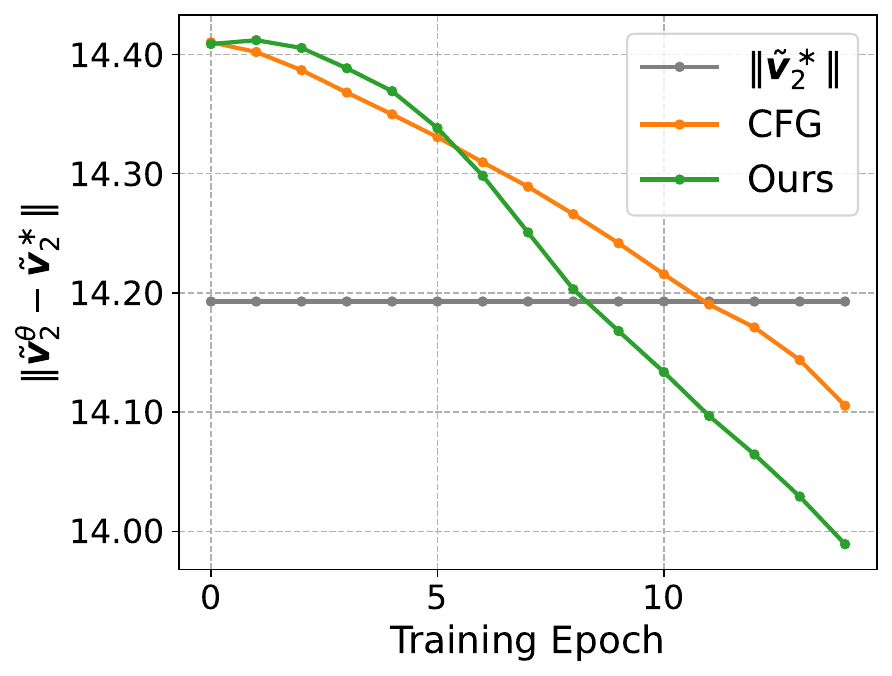} &
    \includegraphics[width=0.25\linewidth,]{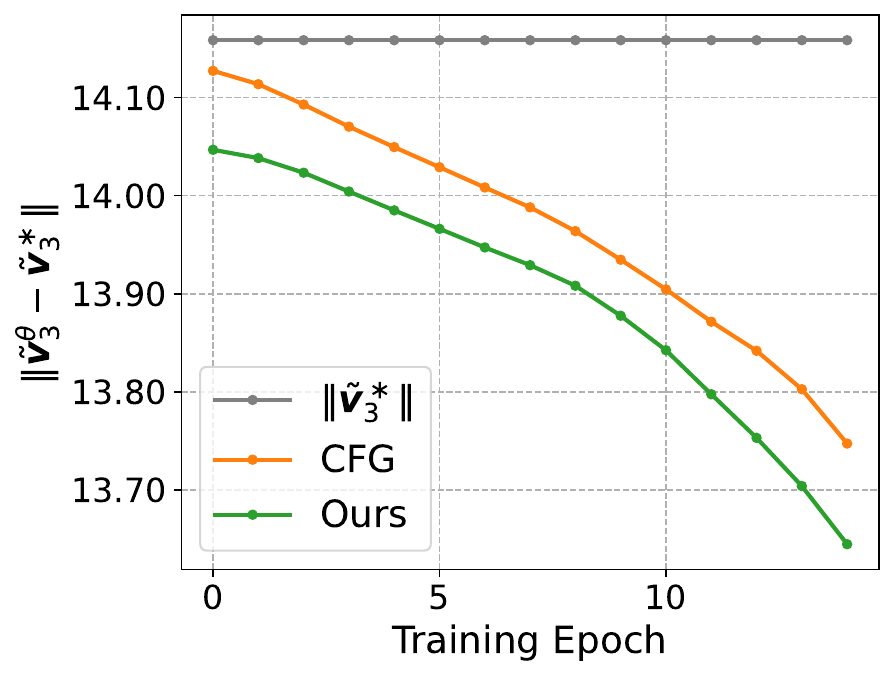} \\
    (a) Zero-init $\tilde{\vv}^\theta_0$ &
    (b) Zero-init $\tilde{\vv}^\theta_0$ and $\tilde{\vv}^\theta_1$ &
    (c) Zero-init $\tilde{\vv}^\theta_0$ to $\tilde{\vv}^\theta_2$ &
    (d) Zero-init $\tilde{\vv}^\theta_0$ to $\tilde{\vv}^\theta_3$ \\
    \end{tabular}
    \caption{Ablation study of zero-init steps.}
    \label{fig:toy_ablation_zero}
\end{figure*}

\noindent\textbf{Ablation study on the number of zero-Initialization steps} is presented in~\figref{fig:toy_ablation_zero}. Specifically, we initialize the first 1, 2, 3, or 4 steps with zeros and observe that during the early training epochs, increasing the number of zero-initialized steps can be beneficial. However, as the number of zero-init steps increases, the learned model achieves better velocity compared to using zero initialization. At this stage, it becomes better to avoid zeroing out additional steps.

\subsection{Experiments on Text-to-Video Generation}
\section{Additional Visual Results}
\label{supp:experiments}
\begin{figure*}[t]
\centering
\includegraphics[width=0.90\linewidth]{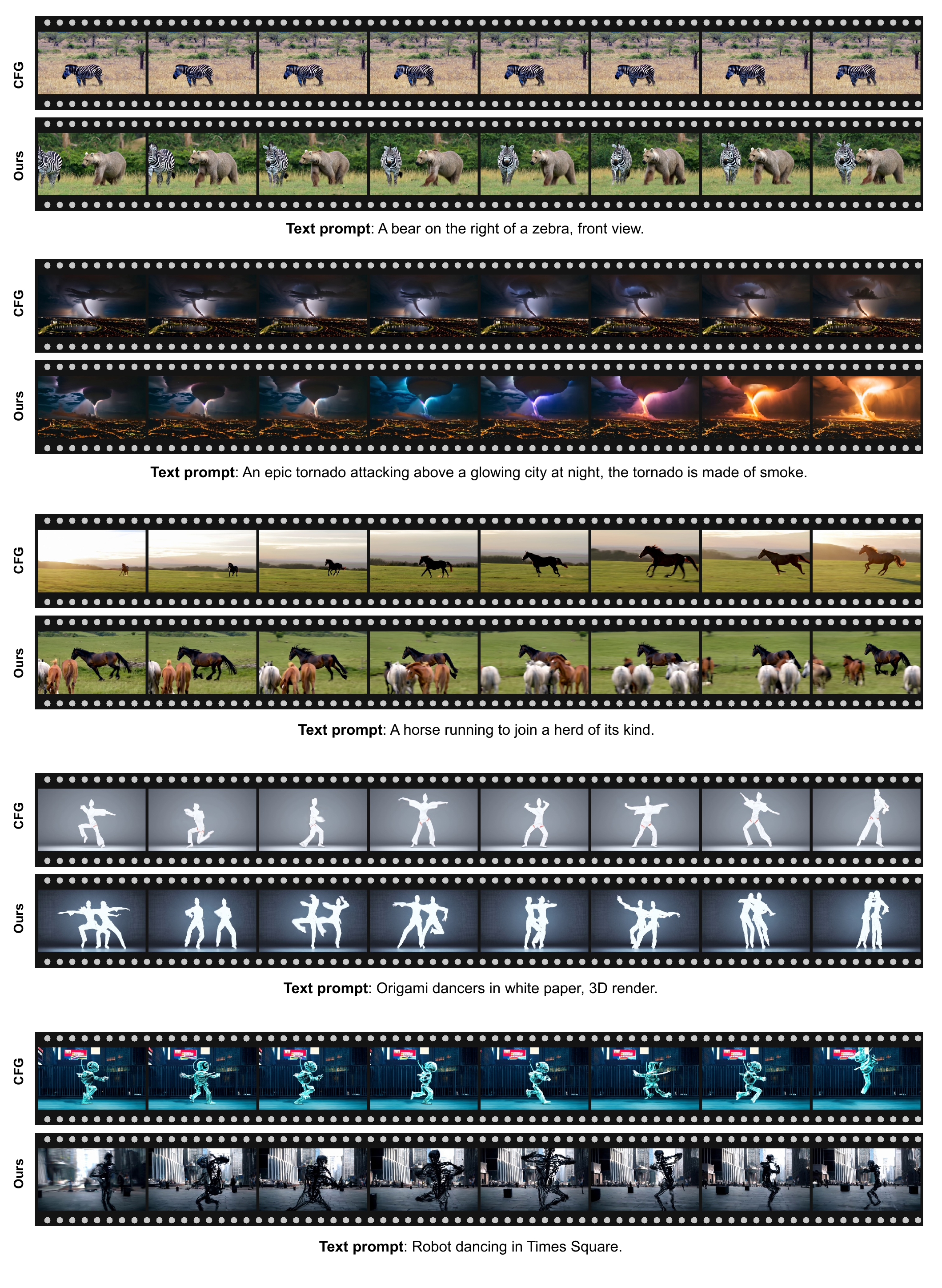}
\caption{\textbf{Additional visual results.} Videos generated by Wan2.1-1B~\cite{wan2.1}}
\label{fig:video_supp_1}
\end{figure*}

\begin{figure*}[t]
\centering
\includegraphics[width=0.90\linewidth]{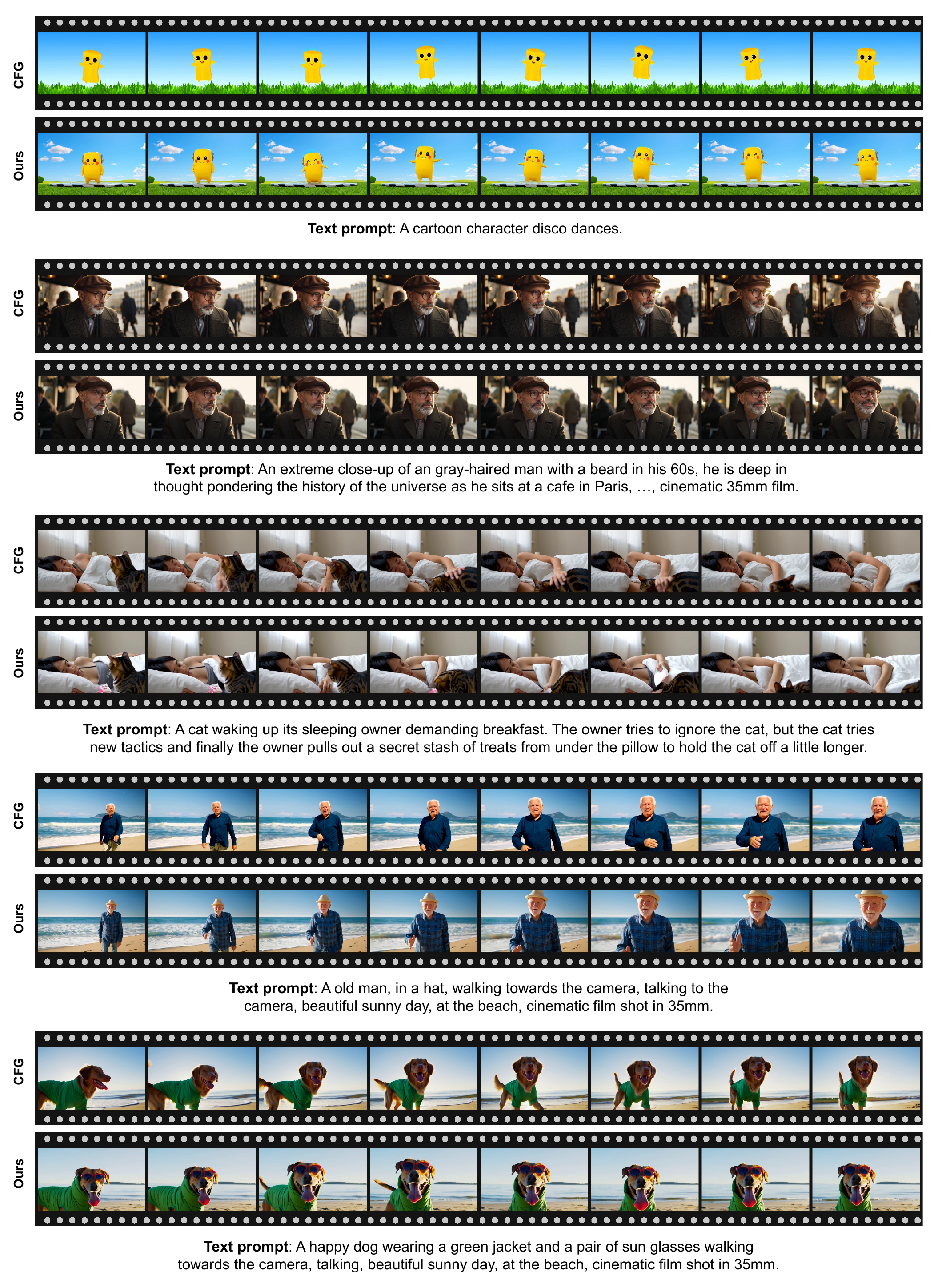}
\caption{\textbf{Additional visual results.} Videos generated by Wan2.1-14B~\cite{wan2.1}}
\label{fig:video_supp_2}
\end{figure*}

\begin{figure*}[t]
\centering
\includegraphics[width=0.90\linewidth]{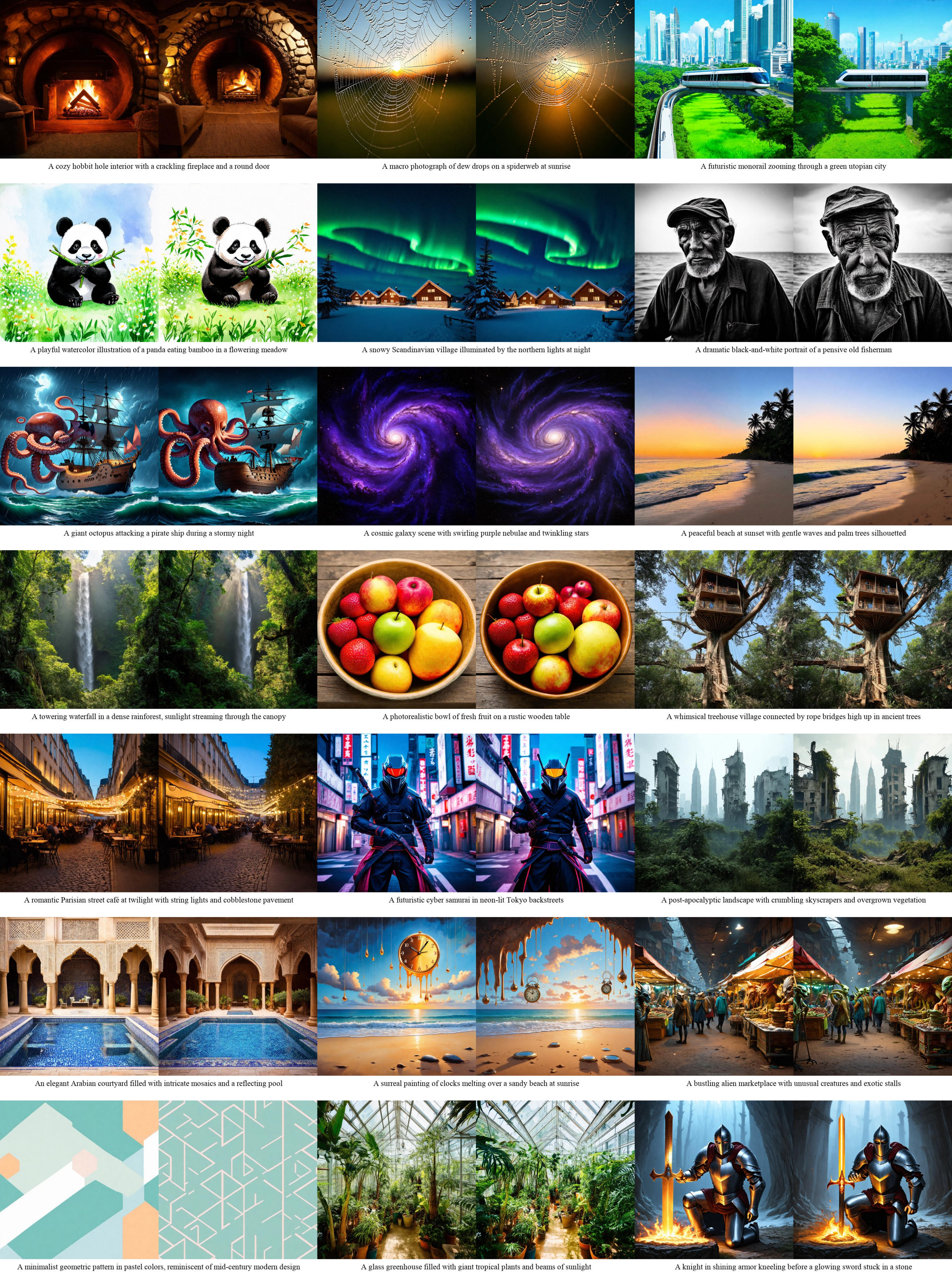}
\caption{\textbf{Additional visual results.} Qualitative comparison between CFG (\textbf{left}) and \ours(\textbf{right}). (Images generated by SD3.)}
\label{fig:ap_img_1}
\end{figure*}

\begin{figure*}[t]
\centering
\includegraphics[width=0.90\linewidth]{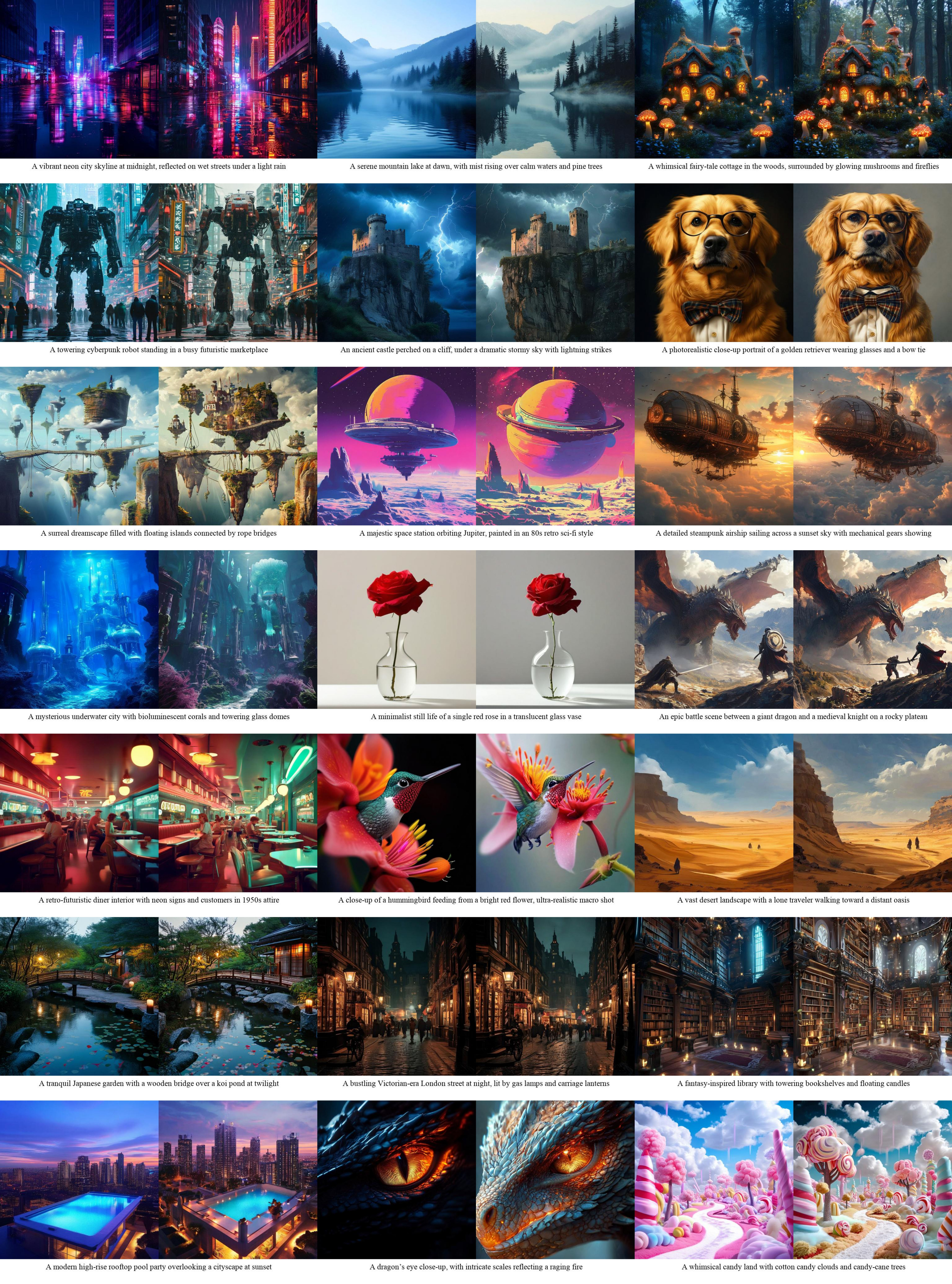}
\caption{\textbf{Additional visual results.} Qualitative comparison between CFG (\textbf{left}) and \ours(\textbf{right}). (Images generated by Lumina-Next.)}
\label{fig:ap_img_0}
\end{figure*}

\begin{figure*}[t]
\centering
\includegraphics[width=0.90\linewidth]{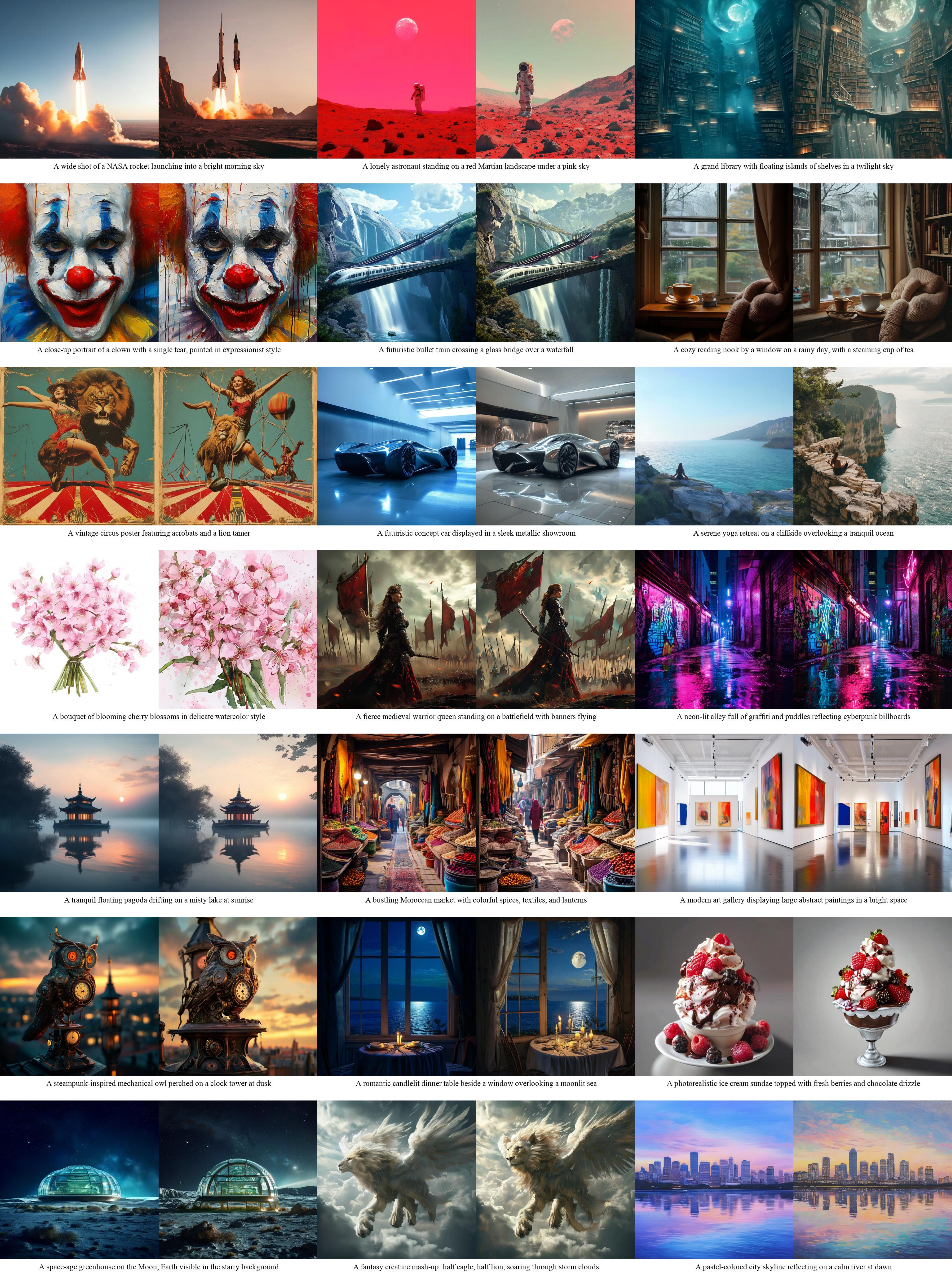}
\caption{\textbf{Additional visual results.} Qualitative comparison between CFG (\textbf{left}) and \ours(\textbf{right}). (Images generated by Lumina-Next.)}
\label{fig:ap_img_3}
\end{figure*}

\begin{figure*}[t]
\centering
\includegraphics[width=0.90\linewidth]{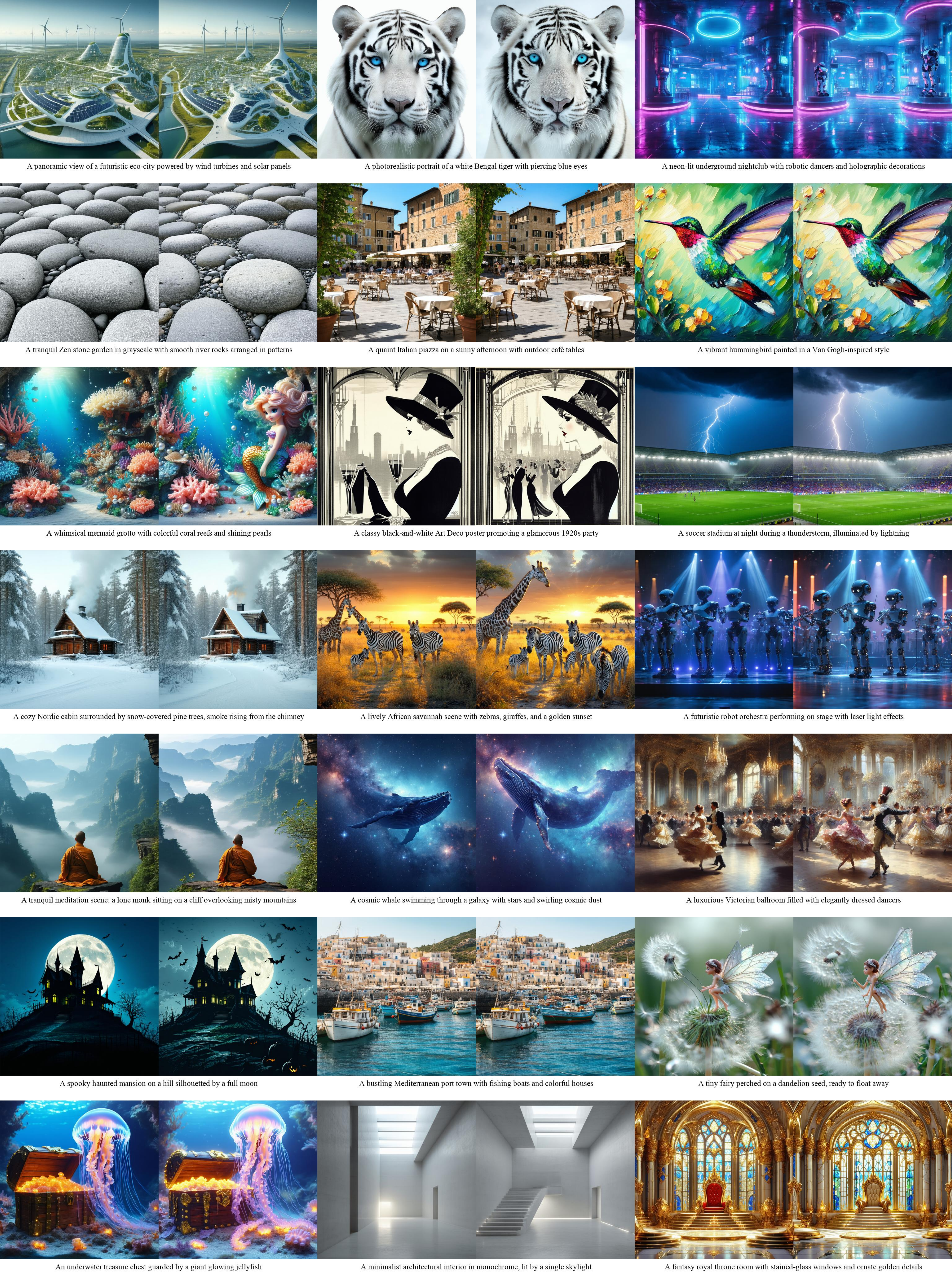}
\caption{\textbf{Additional visual results.} Qualitative comparison between CFG (\textbf{left}) and \ours(\textbf{right}). (Images generated by SD3.5.)}
\label{fig:ap_img_2}
\end{figure*}

In this section, we provide additional visual comparisons between our method and CFG. \figref{fig:video_supp_1} presents videos generated by Wan2.1-1B, while \figref{fig:video_supp_2} showcases those produced by the larger Wan2.1-14B model. Compared to CFG, our videos exhibit finer details, more vibrant colors, and smoother motion. \cref{fig:ap_img_1,fig:ap_img_0,fig:ap_img_2,fig:ap_img_3} provide qualitative comparisons between \ours and CFG. All results are shown without cherry-picking.


\section{Implementation Details}
We first present our code, which can be easily integrated with any Flow-Matching-based model.
\begin{lstlisting}[language=Python]
def optimized_scale(positive_flat, negative_flat):

    # Calculate dot production
    dot_product = torch.sum(positive_flat * negative_flat, dim=1, keepdim=True)

    # Squared norm of uncondition
    squared_norm = torch.sum(negative_flat ** 2, dim=1, keepdim=True) + 1e-8

    # st_star = v_cond^T * v_uncond / ||v_uncond||^2
    st_star = dot_product / squared_norm
    
    return st_star

# Get the velocity prediction
noise_pred_uncond, noise_pred_text = model(...)

positive = noise_pred_text.view(Batchsize,-1)
negative = noise_pred_uncond.view(Batchsize,-1)

# Calculate the optimized scale
st_star = optimized_scale(positive,negative)
# Reshape for broadcasting
st_star = st_star.view(Batchsize, 1, 1, 1)

# Perform CFG-Zero* sampling
if sample_step == 0:
    # Perform zero init
    noise_pred = noise_pred_uncond * 0.
else:
    # Perform optimized scale
    noise_pred = noise_pred_uncond * st_star + \
             guidance_scale * (noise_pred_text - noise_pred_uncond * st_star)
\end{lstlisting}



    





\subsection{Text-to-Image Details}
\noindent\textbf{Quantitative Evaluation.}  
We evaluate Lumina-Next, Stable Diffusion 3, Stable Diffusion 3.5, and De-distill Flux on our self-curated text prompt benchmark, which consists of 200 short and long prompts covering a diverse range of objects, animals, and humans. Each model is assessed using its default optimal settings. For a fair comparison, we generate 10 images per prompt for each model.\\[1ex]
\noindent\textbf{Benchmark Results.}
We compare our method with CFG using Lumina-Next, SD3, and SD3.5 on T2I-CompBench~\cite{huang2025t2icompbench++}, available at \url{https://github.com/Karine-Huang/T2I-CompBench/tree/main}. Each image is generated 10 times with different random seeds to ensure a fair comparison, and all models are evaluated using their optimal settings.\\[1ex]
\noindent\textbf{User Study.}
Our user study includes 76 participants, all familiar with text-to-image generation. Each participant answers a questionnaire consisting of 25 questions, with each question randomly sampled from our generated images in T2I-CompBench~\cite{huang2023t2icompbench,huang2025t2icompbench++} to ensure fair comparisons.

\subsection{Text-to-Video Details}  
We evaluate Wan2.1~\cite{wan2.1} both quantitatively and qualitatively, with all videos generated using the default settings specified in the official repository~\cite{wan2.1} (\url{https://github.com/Wan-Video/Wan2.1}). The VBench evaluation strictly follows the official guidelines (\url{https://github.com/Vchitect/VBench/tree/master}).


\end{document}